\documentclass[11pt, a4paper]{article}

\usepackage[margin=1in]{geometry}
\usepackage{amsmath, amssymb, amsfonts}
\usepackage{graphicx}
\usepackage{booktabs}
\usepackage{multirow}
\usepackage{caption}
\usepackage{subcaption}
\usepackage{siunitx}
\usepackage{hyperref}
\usepackage{authblk}

\usepackage{subcaption}
\usepackage{siunitx}
\usepackage{nameref}

\usepackage{threeparttable}
\usepackage{multirow}
\DeclareMathOperator{\Chenc}{ChEnc}
\DeclareMathOperator{\agg}{Agg}
\DeclareMathOperator{\tenc}{TEnc}
\DeclareMathOperator{\attnlayer}{AttnBlock}
\DeclareMathOperator{\softmax}{softmax}

\newcommand{\numch}{N_{\text{ch}}}
\newcommand{\numt}{N_{\text{t}}}
\newcommand{\demb}{d_{\text{emb}}}

\newcommand{\wei}{Weimann and Conrad}

\newcommand{\code}{\texttt}

\title{Hierarchical Self-Supervised Representation Learning Framework for Multivariate Time Series Grounded in~ECG~Analysis
}

\author{Siwon Kim}
\affil{Research Institute of Basic Sciences, Seoul National University, Seoul, Korea
}
\date{\today}

\begin{document}

\maketitle

\begin{abstract}
    Data analysis in the medical domain often encounters scenarios
involving a limited target dataset and a large, unannotated dataset
with a general distribution.
Under such circumstances, self-supervised learning (SSL) methods are highly effective for utilizing large datasets, making them a popular choice for electrocardiogram (ECG) analysis. 
This work presents the Event Reconstruction Joint-Embedding Predictive Architecture (ER-JEPA), a lightweight SSL framework
for multivariate time series, whose name and two-fold hierarchical structure
are inspired by the diagnostic approach of cardiologists.
At its core, ER-JEPA features: (1)
a two-stage structure that constructs representations for each time interval and subsequently processes these representations as a univariate time series,
(2) the hierarchical integration of two Joint-Embedding Predictive Architectures (JEPAs), and
(3) a Vision Transformer (ViT) backbone.
The structural concatenation of two JEPAs categorizes the model as a Hierarchical JEPA (H-JEPA),
designed to encode multiple levels of abstract representations
for enhanced prediction on complex tasks.
This study reports a successful application
of H-JEPA to 12-lead ECG data as a multivariate time series,
alongside an analysis of the sensitivity of hierarchical representation during the pretraining stage.
Furthermore, this study provides a qualitative demonstration that the intermediate representations produced by the first module of ER-JEPA excel at local feature extraction, as they are structurally free from over-smoothing.
Pretrained on approximately 180,000 10-second recordings, the model achieves state-of-the-art downstream performance on the ST-MEM benchmark, with rapid computation and minimal resource usage.

\end{abstract}

\section{Introduction}\label{sec:introduction} 
While data is the essence of any analysis, from fundamental statistics to deep learning,
the volume of annotated data has rarely scaled to match modern computational demands.
This led to the advent of self-supervised learning (SSL), utilizing learning based on the inherent structure of
data, as seen with invariance-based methods~\cite{chen2020simple,grill2020bootstrap,he2020momentum} and generative methods~\cite{MaskedAutoencoders2021,bao2021beit,xie2022simmim}.
By learning the essence of data through pretraining on large-scale datasets without explicit labels,
SSL frameworks establish a robust foundation for learning a specific task of interest in the downstream fine-tuning step.
Hence, SSL has been extensively utilized in domains with high data availability, such as electrocardiogram (ECG) analysis.

The ECG is a recording of the heart's electrical signals, 
a non-invasive test measuring voltage from electrodes attached to the skin~\cite{thaler2021only}.
Measured from diverse locations, the ECG is a multivariate time series, with \num{12} channels in modern standards.
As one of the most prevalent tests in the clinical field, the ECG yields large-scale datasets within the medical domain.
However, as with most medical data, the ECG requires annotation by professional technicians for precise classification.
Furthermore, due to an abundant number of samples from healthy individuals and typical conditions, the general distribution of ECG data is
heavily skewed, requiring rigorous curation to collect targeted clinical data.
Turning these constraints into an advantage, various studies have explored
the application of SSL frameworks for ECG analysis~\cite{na2024guiding,tian2024foundation,mckeen2025ecg,weimann2025self,kim2024learning}.

Recent research involving SSL frameworks often utilizes the transformer architecture~\cite{vaswani2017attention}
to learn general representations of the input data type, as demonstrated in~\cite{devlin2019bert}.
Frameworks such as Masked Autoencoders (MAE)~\cite{MaskedAutoencoders2021}, which generate masked data, and 
Joint-Embedding Predictive Architectures (JEPA)~\cite{lecun2022path,assran2023self}, 
which predict embeddings in the representation space, are standard building blocks
in recent studies.
This trend is also prominent in ECG analysis; 
however, unlike standard applications in the domains of natural language processing or computer vision,
adaptation for ECG data has not been optimal because it does not share identical dimensional characteristics with either domain.
Following implementations from computer vision, structures designed for two-dimensional data
are not optimal in terms of resource usage.
Some frameworks designed for data types with a linear sequential order are more efficient at processing multichannel time series,
but they do not examine the full potential of each architecture with respect to multichannel processing, since
the multichannel analysis is often handled during the tokenization process.

In recent trends concerning general representation processing,
architectures utilizing hierarchical representations are rising in prominence, particularly in the context of world models~\cite{ha2018world},
which require architectures capable of making inferences based on both fine and coarse semantic understanding~\cite{lecun2022path}.
Incorporating diverse levels of representation is not only crucial for world model learning but is also required for the diagnostic procedure of a cardiologist during ECG analysis.

In this work, we explore the potential of hierarchical representation learning with JEPA on ECG data.
Motivated by the approach of cardiologists in ECG analysis,
this study introduces the Event Reconstruction Joint-Embedding Predictive Architecture (ER-JEPA),
a hierarchical JEPA (H-JEPA) featuring a two-stage structure.
This structural separation allows each module to focus on a designated analytical task, where 
the first part concentrates on multichannel processing and the latter part is dedicated exclusively to temporal analysis.
Moreover, the essence of the structure lies in the concept that the channel module encodes multichannel sequences into
univariate sequences, and the temporal module processes these encodings as a univariate sequence rather than a multivariate sequential input.
By reducing the analysis of multivariate time series into a univariate case,
the model gains significant advantages with respect to efficiency.

Compared to prior studies on the general representation of ECG data using transformer-based SSL frameworks,
ER-JEPA achieves computational efficiency without sacrificing multichannel analysis
by restricting it to a dedicated module.
Another core design choice is the adoption of two separate JEPAs for each part of the model.
While the concatenation of two JEPAs is a double-edged design due to the risk of representation collapse,
empirical results demonstrate that it successfully facilitates richer semantic representation learning.

This report demonstrates the following:
\begin{itemize}
	\item ER-JEPA presents a highly efficient, lightweight solution among ViT models with multichannel processing, achieving a substantial reduction in both memory usage and inference time.
	\item ER-JEPA succeeds in learning hierarchical representations on ECG data without representation collapse, despite utilizing a concatenated JEPA structure
		that is inherently susceptible to the phenomenon.
	\item To mitigate variance in model performance across pretraining, 
		multiple encoders were pretrained to assess optimal strategies and hyperparameters for ER-JEPA
		on ECG data.
	\item ER-JEPA consistently achieves competitive performance with transformer-based 
		SSL models on ECG datasets.
		Specifically, it matches state-of-the-art performance on the ST-MEM~\cite{na2024guiding} benchmark using
		PTB-XL~\cite{wagner2020ptb} and CPSC2018~\cite{liu2018open}, and
		it surpasses state-of-the-art performance on the PTB-XL fine-tuning downstream task
		with an AUC of \num{0.936} and \num{0.943} for multi-label and multi-class evaluation, respectively.
	\item The intermediate representations produced by the channel encoder excel at local feature extraction, as they remain structurally free from token uniformity.
\end{itemize}

\section{Background}\label{sec:background} 
In the field of representation learning, self-supervised learning (SSL) is a highly advantageous framework when large, unlabeled datasets are available.
In the absence of explicit annotations, SSL methods derive supervisory signals directly from 
the inherent structure of the data, for instance, by 
predicting a hidden portion from a given context or enforcing similarity across data augmentations.
In particular, SSL frameworks utilizing transformer architectures to extract generalized representations
have become increasingly prevalent.

\subsection{Self-Attention and Vision Transformer}\label{sub:self_attention_and_vision_transformer} 
Introduced by Vaswani et al.~\cite{vaswani2017attention}, self-attention is a token-wise operation that 
computes mutual correlations among tokens within an input sequence to generate context-aware outputs.
Although initially designed for natural language processing, attention mechanisms are now ubiquitous across various domains,
including computer vision.
The Vision Transformer (ViT)~\cite{dosovitskiy2020image} adapts this architecture for image data by processing non-overlapping image patches as an input sequence.
To retain spatial information, it incorporates two-dimensional positional embeddings, which are vectors added to the tokens to encode their relative positions.
Because the attention operation inherently requires discrete tokens, continuous input data must be discretized prior to processing.

While self-attention can be integrated into broader, more complex structures, its core mechanism relies on the following fundamental operation.
Given $N$ input tokens $\{X_i \mid 1 \leq i \leq N \}$ with an embedding dimension of $\demb$,
we define the query ($Q$), key ($K$), and value ($V$) matrices in $\mathbb{R}^{N \times \demb}$
as learnable linear transformations of the input tokens.
Written in matrix form, the output is:
\begin{displaymath}
	\softmax\left(\frac{QK^{\intercal}}{\sqrt{\demb}}\right)V,
\end{displaymath}
where each output row can be interpreted as a linear combination of value vectors weighted by their mutual correlations:
\begin{displaymath}
	\alpha_{i,1} v_1 + \cdots + \alpha_{i,N} v_N, \quad 1 \leq i \leq N.
\end{displaymath}
Here, $q_i, k_i, v_i$ are the rows of the $Q, K, V$ matrices, respectively, and the attention weight $\alpha_{i,j}$ is derived from the scaled dot-product such that $\alpha_{i,j} \propto \exp(\langle q_i, k_j \rangle / \sqrt{\demb})$.

\paragraph{Notation.}\label{par:notation_} 
We refer to the input tokens processed within a single attention operation as a \textit{sequence}, and
the number of input tokens $N$ as the sequence \textit{length}.
With a slight abuse of notation, we denote the attention operation over a token sequence indexed by $I$ as:
\begin{displaymath}
	\{X_{i}^{(k)} \mid i \in I \} = \attnlayer(\{X_i^{(k-1)} \mid i \in I \}), \quad k \in \mathbb{N}.
\end{displaymath}
This notation accommodates flexible sequence lengths but abstracts away the explicit sequential order of the tokens.
To resolve this, we implicitly assume that positional embeddings have already been added to the input patches to preserve their correct order prior to the operation.
Note that the tokens $\{X_i \}$ are treated as algebraic objects to facilitate the understanding of the operations; however, each
corresponds to a vector of size equal to the embedding dimension $\demb$.

\subsection{Image-Based Joint-Embedding Predictive Architectures}\label{sub:image_based_joint_embedding_predictive_architectures} 
\begin{figure}[htbp]
	\begin{center}
		\captionsetup[subfigure]{justification=centering}
		\begin{subfigure}{0.30\textwidth}
		\includegraphics[width=0.95\textwidth]{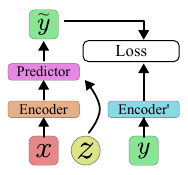}
		\caption{JEPA}\label{subfig:ssl-jepa}
		\end{subfigure}
		\begin{subfigure}{0.5\textwidth}
		\includegraphics[width=0.95\textwidth]{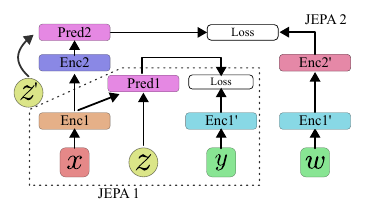}
		\caption{Hierarchical JEPA}\label{subfig:ssl-h-jepa}
		\end{subfigure}
	\end{center}
	\caption{\textbf{Overview of Joint-Embedding Predictive Architecture.}
		(a) The objective of the learning process is the prediction of an embedding from
		a compatible signal utilizing a predictor network, guided by a (possibly latent) variable.
		(b) A basic example of a two-level Hierarchical JEPA.
		The latent-space learning process of JEPA is intrinsically suited for hierarchical composition.
	}\label{fig:jepas}
\end{figure}

Energy-Based Models (EBMs)~\cite{lecun2006tutorial} provide an SSL framework that captures dependencies between variables by learning to assign lower energy to compatible configurations.
The \textit{Joint-Embedding Predictive Architecture} (JEPA)~\cite{lecun2022path} extends this concept using
two encoders to generate representations, followed by a predictor network joining two representations.
Given a pair of intrinsically linked inputs,
JEPA learns their representations by
predicting the output embedding of one encoder from the output embedding of the other encoder, guided by
a latent variable (see Figure~\ref{subfig:ssl-jepa}).

The \textit{Image-based Joint-Embedding Predictive Architecture} (I-JEPA)~\cite{assran2023self} adapts this framework for image
data utilizing a \textit{context} and \textit{target} encoder pair.
It learns the representation of image patches by predicting the embeddings of target patches sampled in a block from
a sparse collection of context patches.
Unlike generative architectures such as MAE~\cite{MaskedAutoencoders2021},
which reconstruct targets in the raw pixel space,
I-JEPA predicts targets exclusively in the abstract representation space.
This design choice is widely cited as a key factor in its enhanced semantic performance.
Consequently, I-JEPA has been successfully adapted for various modalities beyond vision, serving as a core building block for the proposed architecture.

\subsection{Hierarchical JEPA}\label{sub:hierarchical_jepa} 
Each encoder in a JEPA is constrained to designated input specifications, such as a fixed scope and resolution.
For instance, two identical encoders can have distinct input resolutions (e.g., varying sampling frequencies) or scopes (e.g., different patch sizes) 
depending on an associated tokenization.
Furthermore, for multimodal inputs such as an image-text pair,
separate encoders can be assigned to process each distinct modality.

To capture complex data structures, representation learning literature frequently emphasizes the importance of hierarchical processing. In this context,
a \textit{Hierarchical JEPA} (H-JEPA)~\cite{lecun2022path} comprises multiple interconnected JEPAs, where the hierarchy is induced by the varying scopes and resolutions of the sub-encoders.
For instance, a JEPA processing fine representations corresponding to short
time intervals can be chained to another JEPA processing coarser representations
corresponding to longer time intervals (see Figure~\ref{subfig:ssl-h-jepa}).
This aligns with the theoretical premise that complex tasks require multiple levels of 
abstraction, ranging from fine-grained details to coarse semantic concepts, to effectively interpret the input data.

\section{Method}\label{sec:method} 
\begin{figure}[t]
	\begin{center}
		\includegraphics[width=0.95\textwidth]{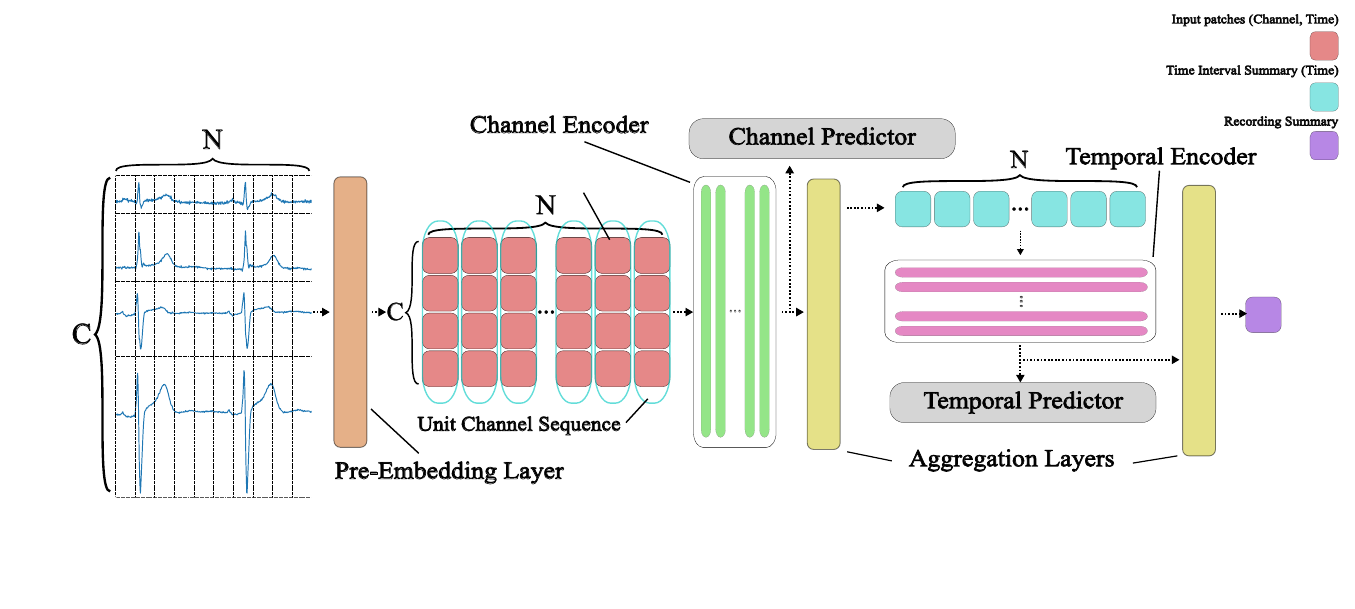}
	\end{center}
	\caption{\textbf{Schematic of ER-JEPA.}
		For a multivariate time series with $C$ channels,
		the pre-embedding layer first tokenizes the recording into $C \times N$ patches indexed by
		(channel, interval). Next, the event reconstruction module, which consists of a channel encoder and
		an aggregation layer, produces a sequence of tokens representing each time interval.
		The event analysis module then processes the resulting sequence to capture temporal relationships across 
		tokens.
		An additional aggregation layer summarizes the interval tokens into a single token,
		which corresponds to a summary of the entire recording.
		The depicted predictors indicate their locations as part of the corresponding JEPAs 
		and do not contribute to inference.
	}\label{fig:schematic-of-er-jepa}
\end{figure}

Given the provided background, the most straightforward interpretation of ER-JEPA is a concatenation of two ViT-based I-JEPAs
operating over different data scopes.
Specifically, the first JEPA, which we refer to as the \textit{channel JEPA}, concentrates on learning inter-channel relationships,
processing tokens by groups of concurrent patches across different channels.
The latter, the \textit{temporal JEPA}, focuses on learning intertemporal dependencies across time intervals.
Functionally, the first module (containing the channel JEPA) summarizes each time interval of a multivariate time series
into a single comprehensive representation. The second module (containing the temporal JEPA)
then processes these constructed representations, focusing exclusively on temporal relations.
We refer to Figure~\ref{fig:schematic-of-er-jepa} for the schematic of the model.

Throughout the discussion, we will consider a multivariate time series as a collection of patches where each patch is a vector representing a specific time interval of some channel, utilizing a natural two-dimensional index (channel, time).

\subsubsection*{The Challenge of Multivariate Time Series in Transformer Frameworks}\label{sec:problem_of_multivariate_time_series_with_transformer_frameworks} 
The two most direct adaptations of I-JEPA for multivariate time series involve adjusting 
image-handling techniques for temporal data.
The first treats multichannel temporal tokens analogously to image patches,
utilizing the (channel, time) indices as the 2D coordinates of each image patch.
The second converts the multivariate time series into a univariate sequence during the preprocessing step,
e.g., by configuring a 1D convolutional layer to accept multiple input channels
in the tokenization step.
Both methods tokenize the input sample into either 2D or 1D patches, which are subsequently processed by the attention layers
as a unified sequence across all intervals and channels.

While these implementations are straightforward,
both present significant drawbacks. 
The first method requires a rather long sequence length for time series processing,
as the structure is adapted directly from the image domain.
Hence, each additional temporal step adds $\numch$ tokens to the sequence for a $\numch$-channel time series,
resulting in a total sequence length of $N = \numch \times \numt$.
Because the time and memory complexity of the standard self-attention algorithm scales quadratically with the sequence length $N$, this $\numch$-fold increase renders rapid analysis computationally infeasible.
Conversely, fusing multichannel information during the preprocessing step circumvents this quadratic bottleneck, as the embedding dimension $\demb$ scales at most linearly in both time and memory complexity.
However, because this method processes multichannel data prior to the attention mechanism,
it deprives the attention layers of the opportunity to capture inter-channel relationships,
failing to fully utilize the model's capabilities.

\begin{figure}[t]
	\begin{center}
		\includegraphics[width=0.95\textwidth]{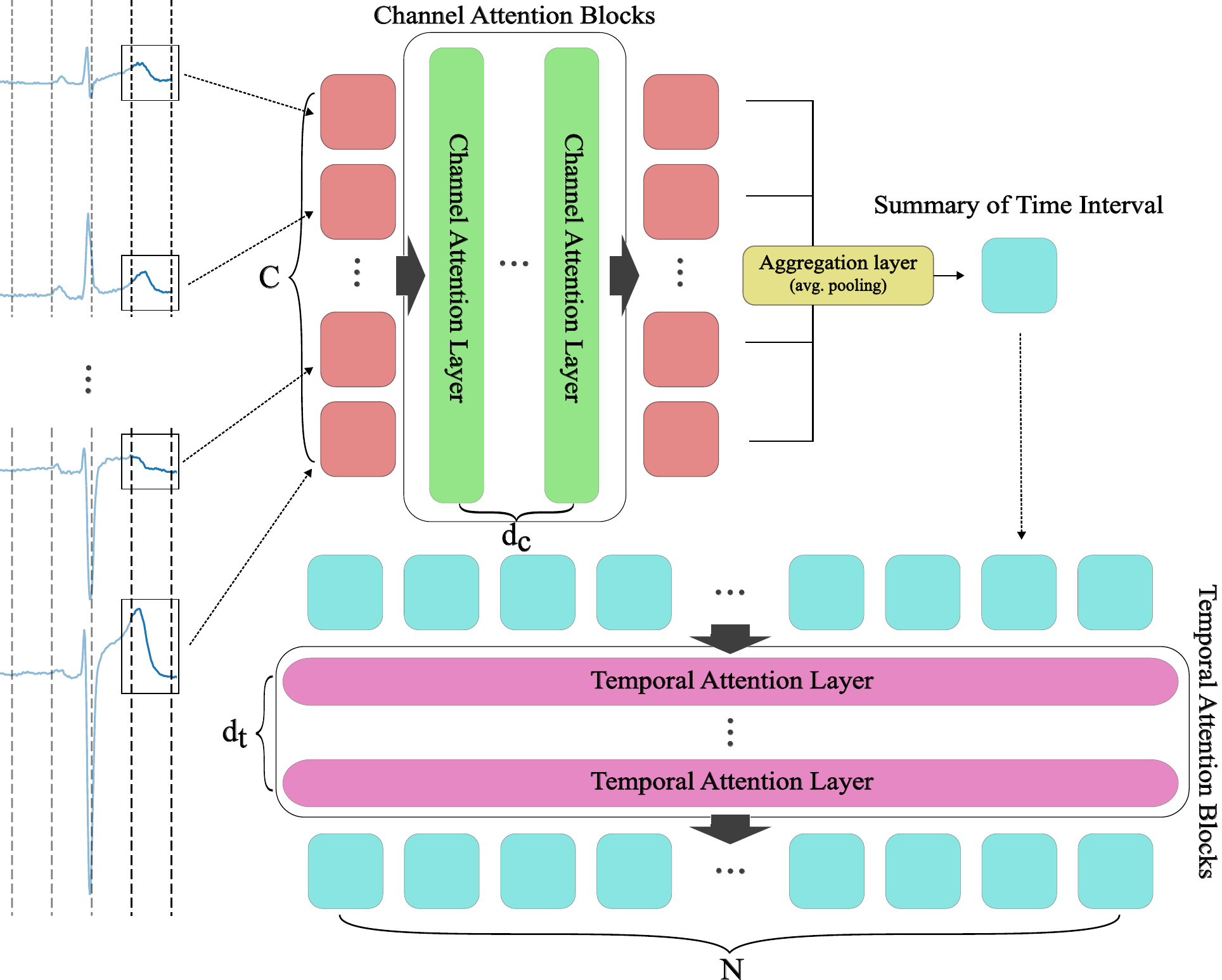}
	\end{center}
	\caption{\textbf{Forward Pass of a Unit Channel Sequence.}
		Given tokens of a multivariate time series indexed by (channel, time interval),
		the channel attention layers process the input by time interval,
		taking the concurrent tokens of every channel as a unit sequence.
		Next, an aggregation layer summarizes the $\numch$ tokens of each time interval into
		a single comprehensive token.
		These summarized tokens then form the input sequence for the temporal attention layer,
		resulting in a sequence length equal to the total number of time intervals.
	}\label{fig:channel-sequence-transform}
\end{figure}

\subsection{Event Reconstruction}\label{sub:event_reconstruction} 
Motivated by the diagnostic approach of cardiologists, ER-JEPA is divided into two modules: (1) the \textit{event reconstruction} module, and (2) the \textit{event analysis} module.
A 12-lead ECG records the electrical activity of the heart projected across 12 spatial axes~\cite{thaler2021only}.
While these projections yield diverse signals, crucially, every projection originates from the \textbf{same underlying cardiac event}. 
Therefore, the diagnostic essence does not lie in
the impulses of a specific channel,
but rather in the holistic inference of the heart's activity derived from all concurrent projections.
This observation suggests that, instead of analyzing patches from every channel,
analyzing a unified representation of the reconstructed cardiac event is a highly efficient strategy in a situation
where multivariate processing can be cumbersome.
Throughout this section, we refer to Figure~\ref{fig:channel-sequence-transform} as a visual guide to
facilitate the understanding of this two-stage token processing.

\paragraph{Event Representation Construction}\label{par:event_representation_construction} 
Given patches of a multichannel time series
\begin{displaymath}
	\{X_{ij}  \mid 1 \leq i \leq \numch, 1\leq j \leq \numt \}
\end{displaymath}
as an input,
the objective of the event reconstruction module is to construct a unified representation for each time interval:
\begin{displaymath}
	\{Y_{j}   \mid 1 \leq j \leq \numt\}.
\end{displaymath}
To yield an informative representation of each time interval, the module must first learn the inter-channel relationships of the given data.
The module achieves this by pairing a channel JEPA with an aggregation layer.
The channel JEPA is a modified I-JEPA dedicated to learning inter-channel relations by
focusing exclusively on the mutual correspondence among
concurrent patches across different channels.
Details regarding the channel JEPA are discussed in Section~\ref{sub:channel_temporal_jepa}.
Through the encoder of the channel JEPA, the input patches are transformed into refined representations of their corresponding channel and interval:
\begin{displaymath}
	\{\tilde{X}_{ij} \}  =  \Chenc(\{X_{ij}\}).
\end{displaymath}
Subsequently, an aggregation layer, such as average pooling, summarizes these concurrent representations into a single, unified representation for that specific interval:
\begin{displaymath}
	\{Y_{j} \} = \agg_i (\{\tilde{X}_{ij} \}  ).
\end{displaymath}

\paragraph{Event Analysis}\label{par:event_analysis_} 
With the constructed representations $\{Y_{j}\}$, the input format for the event analysis module is effectively reduced from multichannel patches to single-channel time patches.
Consequently, the event analysis module is simply a modified I-JEPA model tailored for a univariate time series.
Through the encoder of the temporal JEPA, the constructed representations $\{Y_j\}$ are transformed into a contextually refined embedding of the
corresponding time interval: 
\begin{displaymath}
	\{\tilde{Y}_j\} = \tenc (\{Y_j\}).
\end{displaymath}
This transition from a multivariate time series to a univariate temporal representation is the core driver of the model's efficiency,
reducing the required sequence length for the temporal attention layers.
For downstream tasks requiring a summary of the entire recording, the final temporal outputs are passed through an additional aggregation layer (e.g., average pooling):
\begin{displaymath}
	Z = \agg_j (\{\tilde{Y}_j\})
\end{displaymath}

\subsection{H-JEPA with Channel and Temporal JEPA}\label{sub:channel_temporal_jepa} 
Because the channel JEPA and temporal JEPA are both derived from I-JEPA, their core mechanisms are identical: modules with a ViT backbone; a context encoder, a target encoder, and a predictor for learning joint-embeddings; and updates of the target encoder via an exponential moving average.
For detailed information on I-JEPA, we refer to~\cite{assran2023self}.

\paragraph{Channel JEPA}\label{par:channel_jepa} 
As with the I-JEPA structure, the channel JEPA comprises a context-target encoder pair
and a predictor network, each with a ViT backbone.
And the attention layers inside the channel JEPA receive tokens with two-dimensional indices as input and return processed representations retaining the same index structure.
However, because the primary objective of the channel JEPA is to capture inter-channel relationships, the self-attention operation is applied in a different manner.
Rather than treating all patches across every channel and interval as a massive single sequence,
a unit sequence of the channel JEPA consists of concurrent patches across different channels:
\begin{displaymath}
	\{X_{ij}^{(k)}\}= \bigcup_{j=1}^{\numt} \attnlayer(\{X_{ij}^{(k-1)} \mid 1\leq i \leq \numch \}) 
\end{displaymath}
Processing concurrent patches by each time interval, the model
exclusively captures the mutual correlations among
different channels.
This structural choice reduces the attention sequence length from $\numch \times \numt$ to just $\numch$, shifting the factor of $\numt$ to the total number of self-attention executions.
Accordingly, the context and target masks for the channel JEPA are generated from collections of concurrent channel patches. For any given time interval, the channel JEPA learns by predicting the representations of target channels based on the representations of the provided context channels.

\paragraph{Temporal JEPA}\label{par:temporal_jepa} 
As outlined previously, the temporal JEPA is a direct adaptation of I-JEPA for a single-index sequence, modifying the image-based framework for a univariate time series.
It utilizes the same encoder pair and predictor composition, differing only in its use of 1D positional encoding and its input---which is not derived directly from raw data, but rather from the hierarchical representations constructed by the channel JEPA module:
\begin{displaymath}
	\{Y_{j}^{(k)}\}= \attnlayer(\{Y_{j}^{(k-1)} \mid 1\leq j \leq \numt \}) 
\end{displaymath}
Adjusted for a univariate time series, the context and target masks for the temporal JEPA are sampled from the time intervals.
Hence, the temporal JEPA learns embeddings by predicting representations of target intervals from the provided embeddings of context intervals.

\paragraph{Dual JEPA}\label{par:dual_jepa} 
Note that the predictive learning process of the channel JEPA
is not strictly necessary to form an identical feed-forward structure
to ER-JEPA (see Figure~\ref{fig:schematic-of-er-jepa}).
Since the temporal prediction process occurs at the end of the processing pipeline,
and the ViTs of the channel JEPA (other than the channel encoder) do not participate in the feed-forward process,
it is possible to design an identical model without the channel JEPA.
Omitting the joint-embedding predictive process prior to the temporal encoder
theoretically reduces the risk of representation collapse.
However, as detailed in Section~\ref{par:comparison_without_channel_jepa},
empirical evidence suggests the necessity of the channel JEPA.

\subsection{Adaptation for ECG Data}\label{sub:implementation_for_ecg_data} 

\subsubsection*{Twelve-Lead ECG as a Multivariate Time Series}\label{sec:twelve_lead_ecg_as_multivariate_time_series} 
In standard clinical practice, a 10-second 12-lead ECG recording comprises eight measured leads (I, II, V1, \ldots,~V6),
while the remaining four leads (III, aVL, aVR, aVF) are analytically derived via standard formulas.
Consequently, the adaptation design formulates the 12-lead ECG as an eight-channel time series.
Henceforth, the number of channels is set to $\numch = 8$; and by fixing the \SI{250}{Hz} sampling rate for a \num{10}-second recording, the length of the time series is set to $T = 2500$. 
Formally, the input ECG is represented as a matrix $S \in \mathbb{R}^{8 \times 2500}$ with entries $s_{i,j}$.

\paragraph{Tokenization}\label{par:tokenization} 
Due to the structural differences between image data and time series data,
the patch embedding step is modified for a multivariate time series.
Given a multivariate time series $S \in \mathbb{R}^{\numch \times T}$
and a fixed patch length $p$,
a single patch corresponds to the $j$-th non-overlapping temporal segment of the $i$-th channel, formally denoted as
\begin{displaymath}
S_{ij} = (s_{i,p(j-1)+1}, \ldots, s_{i,p(j-1) + p}) \in \mathbb{R}^{p}.
\end{displaymath}
Consequently, analogous to the image domain, a 1D convolutional layer with both the kernel size and stride set to $p$,
a single input channel, and an output channel equal to the embedding dimension $\demb$ yields the tokenized patch $X_{ij}$ from the time interval $S_{ij}$.

\subsubsection*{Masking Strategy}\label{sec:masking_strategy} 
Inherited from I-JEPA, the two JEPAs require context and target selections for the pretraining process, which are generally referred to as `masks' in the literature.
For each batch, the size of the mask is randomly sampled from a provided range,
and for each sample within the batch, the mask is generated according to the corresponding strategy.
The choice of the selection method depends on the characteristics of the data and the
desired effect on representation learning.

\paragraph{Channel Mask}\label{par:channel_mask} 
Because the pool size for sampling contexts and targets
from the 12 leads is small, a simple random selection is typically preferable.
However, the 12-lead ECG presents an additional constraint that requires a slight modification.
The leads of a 12-lead ECG cover diverse spatial directions:
limb leads (I, II, III, aVL, aVR, aVF) project across vertical directions, 
and precordial leads (V1, \ldots,~V6) project across horizontal directions~\cite{thaler2021only}.
Because only the eight measured leads are utilized in this adaptation, a uniform random selection
would introduce a bias toward the horizontal projections.
To mitigate this bias, the adaptation employs weighted sampling
for mask selection.
Assigning higher sampling weights to leads I and II effectively compensates for the omitted limb leads,
ensuring a balanced selection between the horizontal and vertical spatial directions.

\begin{figure}[htbp]
	\begin{center}
		\captionsetup[subfigure]{justification=centering,skip=0pt}
		\begin{subfigure}{\ifdim\textwidth>14cm 0.85\textwidth \else 0.90\textwidth \fi}
		\includegraphics[width=0.95\textwidth]{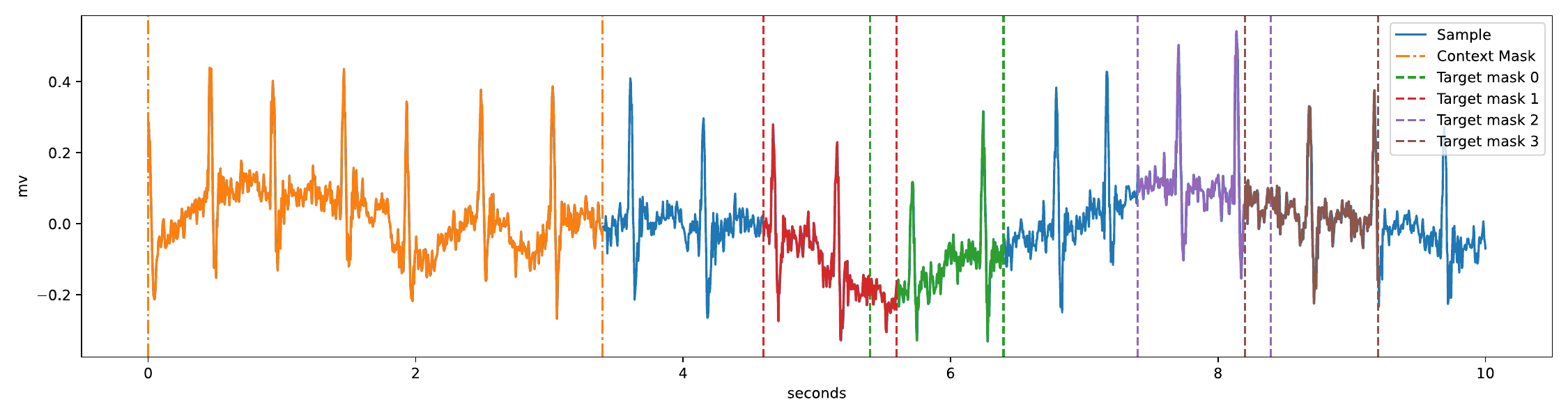}
		\subcaption{Block Mask}
		\end{subfigure}
		\begin{subfigure}{\ifdim\textwidth>14cm 0.85\textwidth \else 0.90\textwidth \fi}
		\includegraphics[width=0.95\textwidth]{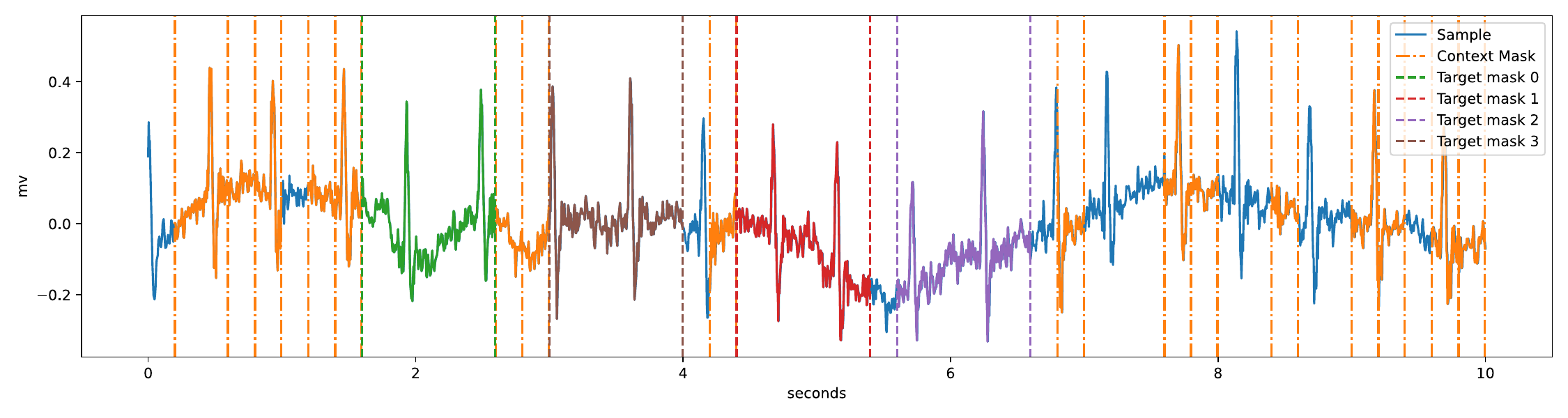}
		\subcaption{Random Mask}
		\end{subfigure}
		\begin{subfigure}{\ifdim\textwidth>14cm 0.35\textwidth \else 0.40\textwidth \fi}
		\includegraphics[width=0.95\textwidth]{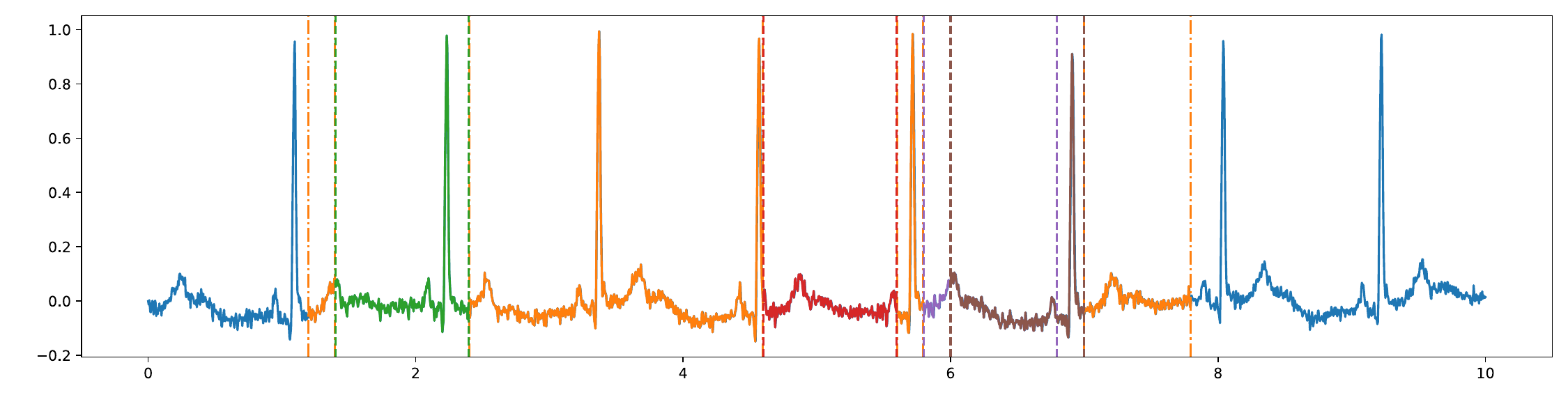}
		\end{subfigure}
		\begin{subfigure}{\ifdim\textwidth>14cm 0.35\textwidth \else 0.40\textwidth \fi}
		\includegraphics[width=0.95\textwidth]{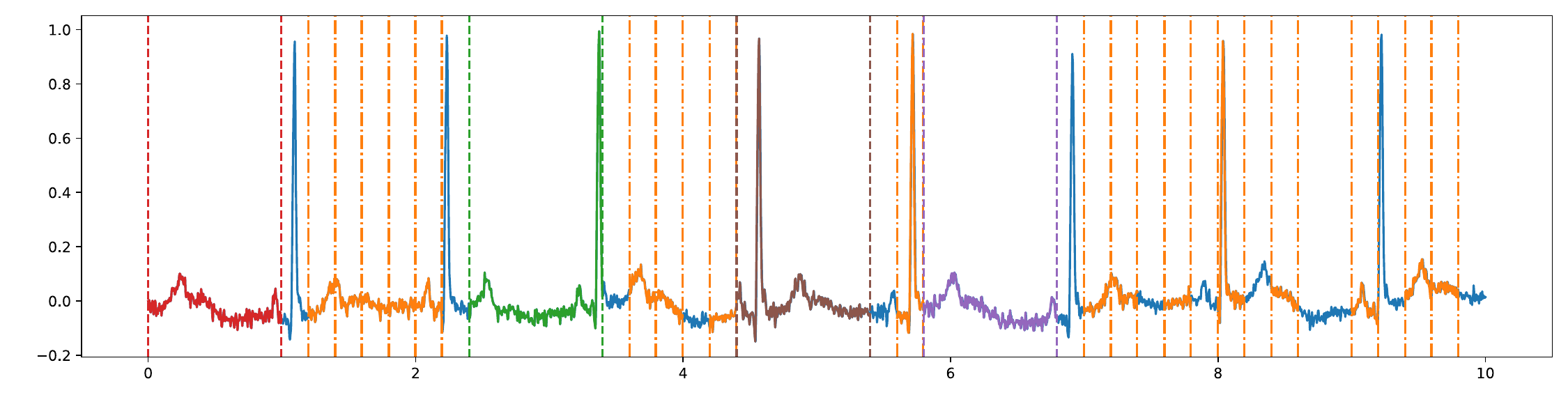}
		\end{subfigure}
		\begin{subfigure}{\ifdim\textwidth>14cm 0.35\textwidth \else 0.40\textwidth \fi}
		\includegraphics[width=0.95\textwidth]{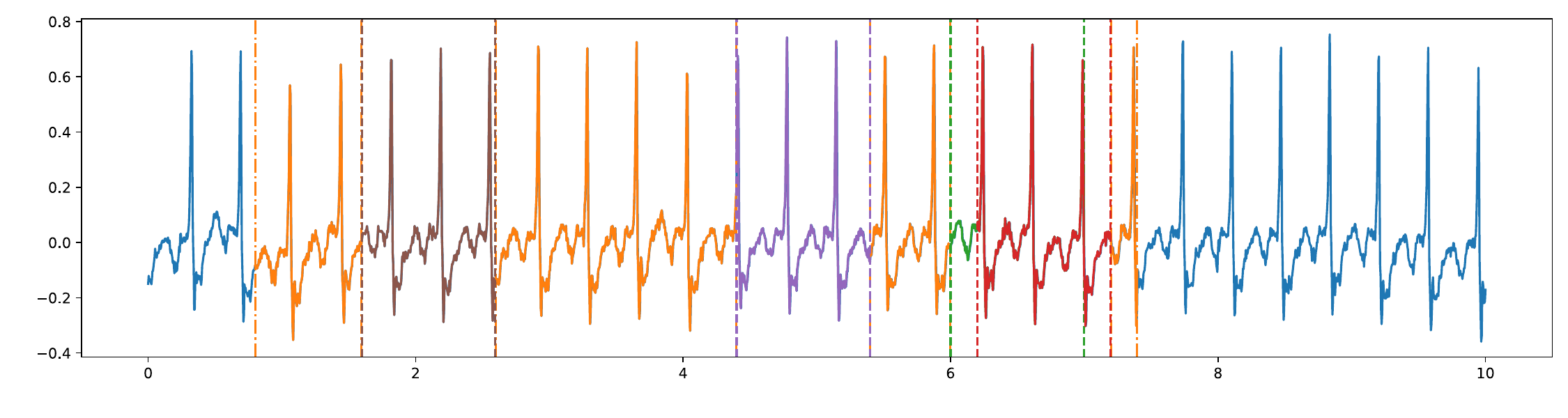}
		\subcaption{Block Masks}
		\end{subfigure}
		\begin{subfigure}{\ifdim\textwidth>14cm 0.35\textwidth \else 0.40\textwidth \fi}
		\includegraphics[width=0.95\textwidth]{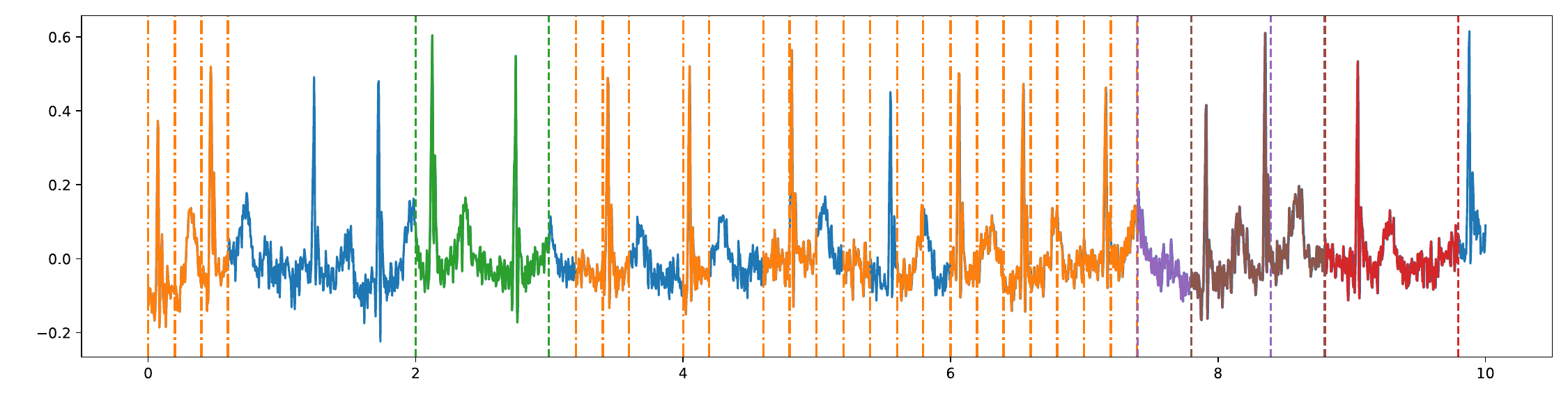}
		\subcaption{Random Masks}
		\end{subfigure}
	\end{center}
	\caption{\textbf{Example of Temporal Masks.}
		Context and target masks for each sample are randomly generated based on a
		predetermined configuration. 
		For each sample within a batch, target masks are sampled as contiguous blocks, with a fixed size across the batch.
		Next, the context mask is sampled from the remaining pool after excluding the target selection;
		the volume of context selection remains constant across the batch, while the sampling pattern depends on the chosen scheme.
		The block mask scheme samples contexts as contiguous blocks (with occasional fragments), the random scheme samples uniformly across the sequence,
		and the mixed scheme applies the block strategy to half of the batch and the random strategy to the remainder.
		For each batch, the context mask size is sampled from the range of \num{1.4}--\SI{3.6}{seconds},
		and the target mask size is sampled from the range of \num{0.8}--\SI{1.4}{seconds}.
	}\label{fig:temporal_mask}
\end{figure}

\paragraph{Temporal Mask}\label{par:temporal_mask} 
Similar to the image domain, target selections are sampled as contiguous blocks to facilitate semantic representation learning.
In this adaptation, two context sampling strategies are evaluated: (1) a block context strategy, and (2) a random context strategy.
Both strategies sample context patches only after excluding the selected target patches from the available pool.
The block strategy aggregates consecutive patches, occasionally fragmented by the excluded target regions, whereas the random strategy uniformly samples individual patches across the remaining sequence.
Figure~\ref{fig:temporal_mask} provides visual examples of each scheme.
For a performance comparison of the two strategies, refer to Section~\ref{par:masking_strategy} and Table~\ref{tab:mask_strategy}.

\section{Related Work}\label{sec:related_work} 
Given the high availability of ECG data, various studies have explored general representation learning for ECG data using SSL.
The work on Spatio-Temporal Masked Electrocardiogram Modeling (ST-MEM) is one of the pioneering works based on the Masked Autoencoder (MAE),
which emphasized the importance of multichannel token processing, referring to patchification with a two-dimensional index as `spatio-temporal patchifying'.
The benchmark methodology presented in their work 
featured downstream classification tasks on the PTB-XL~\cite{wagner2020ptb} and CPSC2018~\cite{liu2018open} datasets,
and it has since been widely adopted in the SSL-based analysis of ECG data.

Additionally, several studies have investigated the application of JEPA to the ECG domain.
For instance, Weimann and Conrad~\cite{weimann2025self} 
explored the large-scale pretraining of I-JEPA, incorporating 
several image-to-time modifications specifically designed for ECG data.
By utilizing a convolutional layer to tokenize the 12-lead ECG into 1D patches,
their model focuses primarily on processing temporal tokens.

Similarly derived from I-JEPA, ECG-JEPA~\cite{kim2024learning} is another predictive framework designed for 12-lead ECGs.
As with ST-MEM, the design of ECG-JEPA accounts for inter-channel relationships through an attention mechanism.
It adopts a structure similar to standard I-JEPA, processing multichannel tokens utilizing a two-dimensional (channel, time) index;
however, it differs by explicitly restricting mutual token influences during the attention operation.
Its Cross-Pattern Attention layer is
based on the standard ViT attention layer for image data,
but for any given target patch, it restricts attention exclusively to patches within
the same channel (horizontal) and concurrent patches across different channels (vertical),
effectively forming a spatial-temporal cross that intersects at the target patch.

Furthermore, the evaluation benchmark of ECG-JEPA follows the methodology established by ST-MEM,
incorporating comprehensive experimental results for baseline models alongside configuration details for reproducibility.
To facilitate a rigorous and direct comparison, the evaluation in this study follows the benchmarking methodology of ECG-JEPA.


\section{Experiments}\label{sec:experiments} 

\subsection{Pretraining}\label{sub:pretraining} 
The paradigm of two-stage training, which first learns the underlying structure of the data using a large unlabeled dataset, is the essence of SSL.
Following the methodology of I-JEPA~\cite{assran2023self}, ER-JEPA learns by predicting the encoding of target patches from context patches during the pretraining process.
Learning robust semantic representations from the inherent structure of the data
fundamentally requires a large-scale dataset.

\subsubsection*{Pretraining Dataset}\label{sec:pretraining_dataset} 
Due to the high availability of ECG data, there are various large-scale 12-lead ECG datasets, notably the \textbf{Chapman-Shaoxing} dataset~\cite{zheng2022large}, comprising \num{45152} recordings, 
and the \textbf{CODE-15} dataset~\cite{ribeiro2021code}, comprising \num{345779} recordings.
Aimed at promoting the analysis of arrhythmias and cardiovascular diseases, the Shaoxing dataset features diverse arrhythmias with labels provided by professional experts, comprising recordings collected from Shaoxing People's Hospital and Ningbo First Hospital.
As a subcollection of the comprehensive CODE dataset~\cite{ribeiro2019tele}, the CODE-15 dataset represents a \SI{15}{\percent} stratified sample of the entire dataset.
Collected by the Telehealth Network of Minas Gerais, this dataset covers a broad patient demographic across Brazil.

\paragraph{Filtering and Preprocessing}\label{par:filtering_and_preprocessing} 
Restricting the scope to \num{10}-second recordings reduces the number of usable recordings to \num{143328} for the CODE-15 dataset.
Additionally, recordings containing NaN values or missing leads were removed,
yielding a total pretraining dataset of \num{174461} recordings (\num{43561} from the Shaoxing dataset and \num{130900} from the CODE-15 dataset).
All recordings were resampled to \SI{250}{Hz}.
No further preprocessing was applied, except for data augmentations utilized to prevent
overfitting across repeated epochs and to enhance the robustness of the learned encodings.

\subsubsection*{Training}\label{sec:training} 
During pretraining, the model was trained over \num{300} epochs with a batch size of \num{64}.
The AdamW optimizer~\cite{loshchilov2017decoupled} was utilized,
adjusted by a Stochastic Gradient Descent with Warm Restarts (SGDR) scheduler~\cite{loshchilov2016sgdr}, 
featuring a base learning rate of \num{2e-4}, a minimum learning rate of \num{5e-5}, eight restarts, 
and \SI{10}{\percent} warmup epochs.
The target encoder was updated via an exponential moving average 
with a linearly increasing momentum value from \num{0.996} to \num{1}, following the configuration of I-JEPA~\cite{assran2023self}.
Training on a workstation equipped with an NVIDIA GeForce RTX 3090 GPU (\SI{24}{GB} VRAM) takes approximately \num{34} hours and requires \SI{6}{GB} of VRAM, or \num{40} hours and \SI{3}{GB} without Just-In-Time (JIT) compilation.

\subsection{Sensitivity of Hierarchical Representation}\label{sub:sensitivity_of_hierarchical_representation} 
Because the learning process of JEPA operates in the representation space,
the architecture is prone to \textit{representation collapse}, 
a phenomenon where the model trivially solves the prediction task by collapsing all encodings into a constant embedding.
This can be largely mitigated by design choices, such as utilizing asymmetric encoders~\cite{baevski2022data2vec,chen2020exploring,grill2020bootstrap}, 
as seen in many Joint-Embedding Architecture (JEA) models and I-JEPA;
however, an inherent susceptibility remains prevalent across joint-embedding architectures.

\begin{figure}[htbp]
	\begin{center}
		\captionsetup[subfigure]{justification=centering}
		\begin{subfigure}{0.95\textwidth}
		\includegraphics[width=\textwidth]{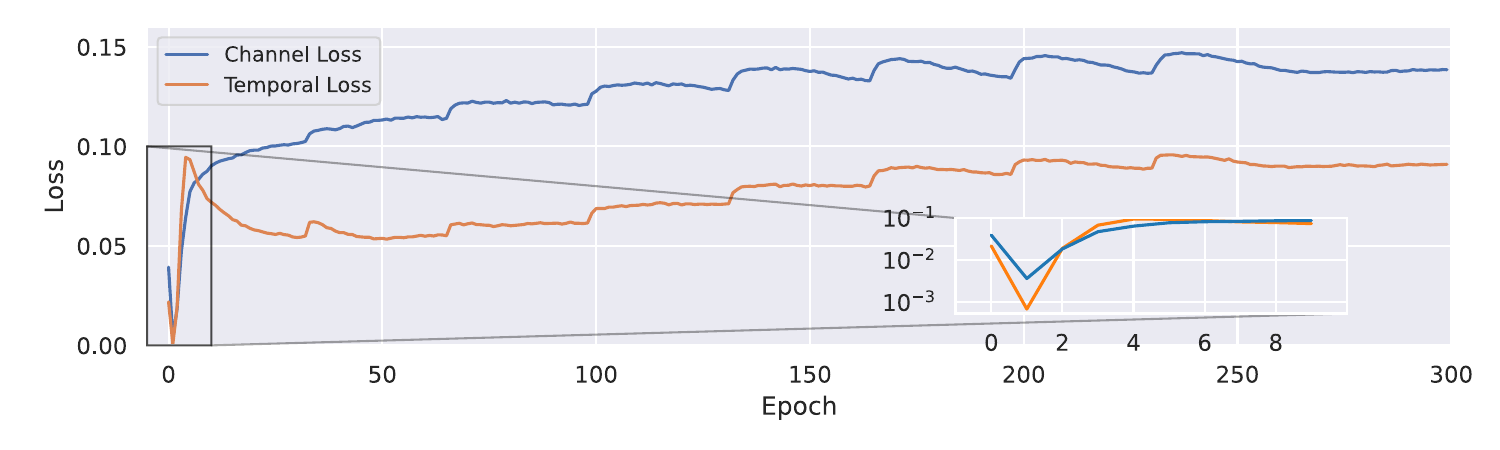}
		\caption{}\label{subfig:pretraining-loss-plot}
		\end{subfigure}
		\begin{subfigure}{0.95\textwidth}
		\includegraphics[width=\textwidth]{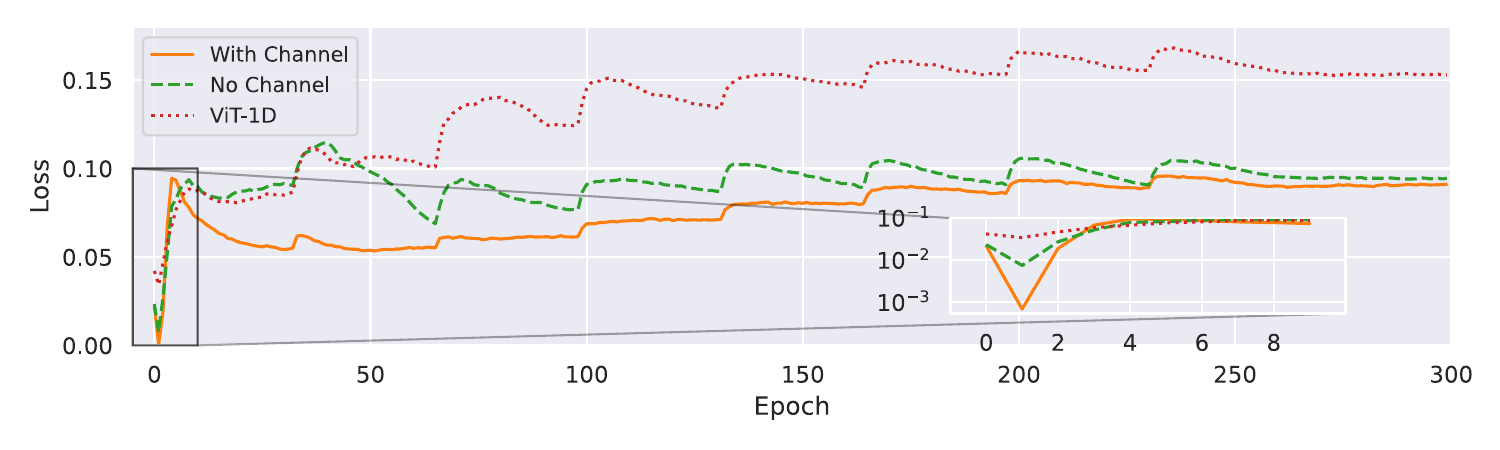}
		\caption{}\label{subfig:pretraining-loss-versus-no-channel}
		\end{subfigure}
	\end{center}
	\caption{\textbf{Pretraining Loss of the Channel and Temporal JEPA.}
		(a) Evolution of channel and temporal loss during pretraining.
		Both JEPAs exhibit an early-stage drop in loss,
		with the temporal loss reaching a lower minimum loss of \num{7e-4}.
		(b) Impact of the channel JEPA on temporal loss.
		The plot compares the temporal loss of the complete model against baseline architectures without a concatenated JEPA structure, including an identical model
		omitting the channel JEPA and a standard 1D I-JEPA adaptation for ECG.
		The reported loss value is the mean of an element-wise comparison.
	}\label{fig:loss}
\end{figure}

Representation collapse is an especially serious issue for ER-JEPA, since the input to the temporal JEPA
is not a token derived directly from the raw data, but rather a representation constructed by the preceding channel JEPA,
which is itself inherently prone to representation collapse.
Indeed, as depicted in Figure~\ref{subfig:pretraining-loss-plot}, the loss of each JEPA drastically drops in the early stages, then recovers within a few epochs, a recovery pattern typical for models utilizing the JEPA framework.

To further investigate this behavior, the evolution of the temporal loss can be compared against baseline models that lack a concatenated JEPA structure, as shown in Figure~\ref{subfig:pretraining-loss-versus-no-channel}.
Under an identical pretraining environment, both an ablation model omitting the channel JEPA (detailed below) and a standard 1D I-JEPA adaptation for ECG (see Section~\ref{sec:benchmark_methodology}) exhibit a significantly shallower initial drop in loss during the early epochs.
This distinction highlights that the compound hierarchical stacking in the complete ER-JEPA model amplifies early-stage vulnerability prior to its recovery.

While the complete model ultimately stabilizes, an unrecovered loss during pretraining serves as typical evidence of representation collapse.
Building on this observation, the following paragraphs detail specific phenomena regarding the sensitive nature of 
hierarchical representation learning within ER-JEPA.

\begin{figure}[htbp]
	\begin{center}
		\captionsetup[subfigure]{justification=centering}
		\begin{subfigure}{0.45\textwidth}
		\includegraphics[width=\textwidth]{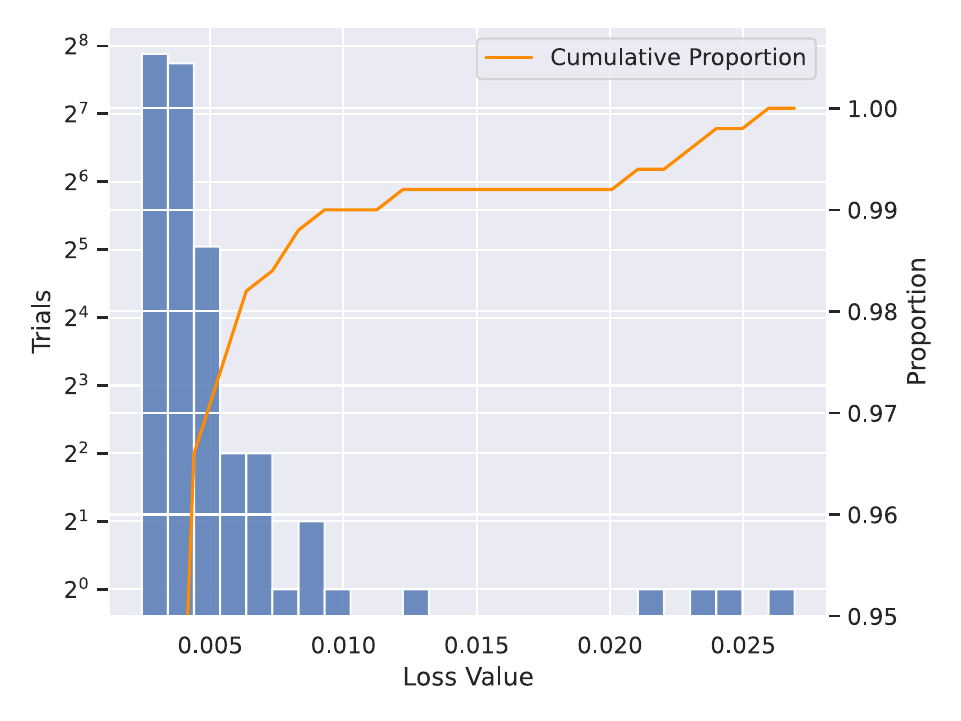}
		\caption{Total Loss}\label{subfig:total-loss-hist}
		\end{subfigure}
		\begin{subfigure}{0.45\textwidth}
		\includegraphics[width=\textwidth]{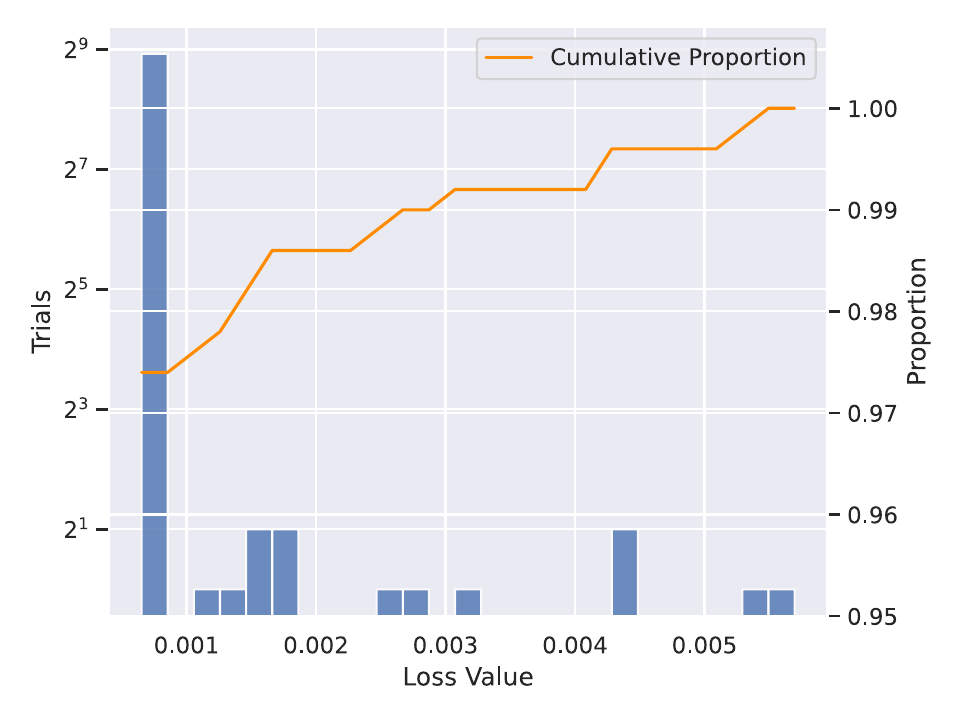}
		\caption{Temporal Loss}\label{subfig:temporal-loss-hist}
		\end{subfigure}
	\end{center}
	\caption{\textbf{Histograms of Loss Value at Epoch 2.}
		Under fixed pretraining configurations, \num{500} repeated trials yielded approximately
		\SI{1}{\percent} anomalous cases, characterized by relatively high loss values at Epoch 2.
		This phenomenon was also consistently observed in the rare full trials that produced 
		inferior downstream performance. The five trials with the highest total loss 
		corresponded to cases of highest loss in both the channel and temporal modules.
	}\label{fig:loss_hist}
\end{figure}

\paragraph{Loss Drop}\label{par:loss_drop} 
Because unsuccessful JEPA designs often do not recover from an infinitesimal loss in the early stages,
a very small initial loss value is typically considered a harbinger of failure.
However, empirical results indicated otherwise for ER-JEPA under specific conditions.
With the current pretraining data and a batch size of \num{64},
repeated trials occasionally produced anomalous pretraining results, yielding inferior downstream performance that significantly deviated from the norm (see Table~\ref{tab:misc}).
While a definitive explanation remains elusive, the degree of the initial loss drop for these anomalous trials was notably \textbf{less} severe than in standard, successful trials.
To provide a clearer view of this phenomenon, the first two epochs of pretraining were sampled across numerous trials
without continuing to the final epoch (see Figure~\ref{fig:loss_hist}).
Approximately \SI{1}{\percent} of the \num{500} trials exhibited significant deviations in the minimum loss value, observable in both the channel and temporal loss.

\paragraph{Comparison Without Channel JEPA}\label{par:comparison_without_channel_jepa} 
An encoder with an identical structure can be trained without the channel JEPA by omitting the predictive process for the channel module.
Because the temporal JEPA remains at the end of the encoding layers, the pretraining process is still valid, and
the risk of representation collapse is theoretically reduced to the level of a single JEPA.
Figure~\ref{subfig:pretraining-loss-versus-no-channel} compares the temporal loss
between the ablation model without the channel JEPA and the complete model.
Even under identical training settings, the removal of the channel JEPA prevents the drastic drop in loss during the early stages.
However, this structural absence results in noticeable degradation in downstream task
performance compared to the complete model (see Table~\ref{tab:misc}).

\paragraph{Dropout Layer}\label{par:dropout_layer} 
In conjunction with the hypothesis regarding early-stage loss drops and the predictive nature of pretraining,
the performance increase observed from utilizing a dropout layer~\cite{hinton2012improving} 
demonstrates a potential relationship 
with sensitive representation learning.
As presented in Table~\ref{tab:misc}, ablating the dropout layer results in a notable decline in downstream task performance,
suggesting that enforcing representational robustness is critical in hierarchical JEPA frameworks.

\subsection{Downstream Tasks}\label{sub:downstream_task} 
The evaluation of representation learning assesses the encoder's performance on 
specific tasks, both with and without further altering the model's weights.
To evaluate the static embeddings produced by the pretrained encoder, \textbf{linear probing} measures the performance of a 
basic linear classifier using the frozen encoder's output as features.
To evaluate the full capacity of the model, \textbf{fine-tuning} 
trains the entire model end-to-end after appending a classification layer.

Following I-JEPA~\cite{assran2023self}, the target encoder of the JEPA framework, which is updated via an exponential moving average, is the standard choice for the \textbf{representing encoder} in downstream applications.
For ER-JEPA, the concatenation of the channel and temporal JEPA encoders, 
joined by an average pooling layer, serves as the primary encoder for these tasks.
For the classification of an entire \num{10}-second \num{12}-lead ECG, an additional average pooling layer is appended to summarize the sequence of interval representations into a single global representation (see Figure~\ref{fig:schematic-of-er-jepa}).

\subsubsection*{Downstream Dataset: Classification}\label{sec:downstream_dataset_classification} 
Following the benchmarking methodology established by ST-MEM~\cite{na2024guiding},
the PTB-XL~\cite{wagner2020ptb} and CPSC2018~\cite{liu2018open} datasets were utilized for the downstream classification tasks.
For direct comparison across literature, the macro area under the receiver operating characteristic curve (macro AUC)
was utilized as the primary metric for all classification tasks.

\paragraph{PTB-XL}\label{par:ptb_xl} 
Structured by the Physikalisch-Technische Bundesanstalt (PTB) in Germany, the dataset~\cite{wagner2020ptb}
comprises \num{21799} clinical \num{10}-second $12$-lead ECG recordings collected from \num{18869} patients.
The annotations, provided by two cardiologists, cover \num{71} different ECG statements conforming to the SCP-ECG standard which are categorized into diagnostic, form, and rhythm statements.
The dataset provides recommended splits for training and testing, alongside
two types of diagnostic labels based on \num{41} diagnostic statements, aggregated into \num{24} \textbf{subclasses} and 5 coarser \textbf{superclasses}.

\paragraph{CPSC 2018}\label{par:cpsc_} 
The China Physiological Signal Challenge (CPSC) 2018 dataset~\cite{liu2018open} comprises a collection of ECG recordings donated by \num{11} hospitals
for the 2018 competition.
The dataset features eight arrhythmia classes across
\num{6877} $12$-lead ECG recordings for training and \num{2954} withheld recordings for testing.

\paragraph{Labels, Training, and Test Sets}\label{par:training_and_testset} 
For PTB-XL, due to its ubiquitous use in related works, 
the superclass was utilized as the target label for classification.
Following the provided stratified splits, folds \num{1} through \num{8} were used as the training set, while fold \num{9} and fold \num{10} served as the validation and test sets, respectively.

For CPSC2018, all \num{9} labels, including the normal sinus rhythm (healthy individual), were used as target classes.
Split by the thousands digit of the record ID, the dataset was divided into \num{7} batches,
with \num{878} recordings in the final batch.
After excluding records with NaN values or missing leads, 
the first five batches were used as the training set, the sixth as the validation set, and the final batch as the test set.
These specific splits for both datasets follow the ECG-JEPA~\cite{kim2024learning} methodology to ensure direct comparability.

\paragraph{Multi-Label and Multi-Class Tasks}\label{par:multi_label_and_multi_class_tasks} 
Depending on the label structure, the evaluations were further divided into two downstream configurations: multi-label classification and multi-class classification. 
In the multi-label classification task, multiple labels for a single sample are permitted, 
and the dataset is used without filtering. 
Conversely, in the multi-class setting,
samples with multiple labels are filtered out, retaining only those with a single definitive label.

\subsubsection{Benchmark Methodology}\label{sec:benchmark_methodology} 
Table~\ref{tab:linear_eval} and Table~\ref{tab:finetune} present the corresponding benchmark results of SSL models on ECG data.
The reported scores of SSL models from the original ST-MEM~\cite{na2024guiding} paper
are listed as a baseline,
along with the test results of the JEPA model from Weimann and Conrad~\cite{weimann2025self},
noting possible differences in their splitting of the CPSC2018 dataset.
The scores for SimCLR~\cite{chen2020simple}, ECG-FM~\cite{mckeen2025ecg}, 
KED~\cite{tian2024foundation}, and ECG-JEPA 
are derived from the comparative tests in the work introducing ECG-JEPA~\cite{kim2024learning}.
To utilize  this extensive evaluation, the assessment of ER-JEPA strictly follows the ECG-JEPA setup for direct comparison.

As an additional baseline, a simple 1D-JEPA was trained within an identical pretraining environment to ER-JEPA.
The structure of this 1D-JEPA follows the framework described in Section~\ref{sec:problem_of_multivariate_time_series_with_transformer_frameworks},
where multichannel processing is handled entirely by a convolutional pre-embedding step.
This baseline is conceptually identical to the JEPA framework of \wei, though specific architectural details may differ.
Furthermore, the \wei{} JEPA operates on a higher-resolution signal input (patch size \num{25}, \SI{500}{Hz} sampling rate)
and utilizes a vastly different data-intensive pretraining regime (sampling \SI{1}{M} recordings from
\num{10} public datasets, batch size \num{2048},
training for up to \num{100}K epochs, with downstream evaluations starting from a minimum of \num{10}K epochs).
Therefore, considering these differences in data resolution and pretraining scale, the locally trained 1D-JEPA serves as a more controlled surrogate for evaluating the architectural benefits of ER-JEPA.

\paragraph{Downstream Training}\label{par:downstream_training} 
The downstream evaluation procedures are outlined below.
For the classification task, 
a linear layer, prepended with batch normalization, is appended to serve as the classification head.
Taking the pooled representation of the entire recording as input (see Figure~\ref{fig:schematic-of-er-jepa}), the classification layer predicts the diagnostic label.
Each evaluation process trains the model on the training set, utilizes early stopping based on the validation set, and reports the final metrics from the test set.
Unless otherwise specified, every repeated experiment features a distinctly pretrained encoder and experiments were repeated ten times for the benchmark comparison and five times for ablation studies.

\subsubsection{Linear Evaluation}\label{sec:linear_evaluation} 
The encoder of an SSL model learns embeddings that capture the essence of the data
by learning inherent data structures during pretraining.
Linear probing evaluates the quality of these embeddings by freezing the encoder's weights 
and only training the appended linear classifier.
For details regarding the linear classifier and the evaluation process, refer to the \nameref{par:downstream_training} paragraph above.
The linear classifier was trained over \num{70} epochs using a decreasing learning rate 
starting from \num{5e-3}, a batch size of \num{16}, and the AdamW optimizer~\cite{loshchilov2017decoupled}.
\begin{table}[ht]
	\centering
	\setlength{\tabcolsep}{4.5pt}
	\caption{\textbf{Linear Evaluation Benchmark Results.}
		Performance (macro AUC) of SSL models on downstream datasets with linear evaluation.
		Entries with multiple trials are accompanied by the standard deviation in parentheses
		(0.0xx), and the reported value is the average macro AUC over trials.
		The \textit{Source} column indicates the literature from which the benchmark scores were retrieved.
	}\label{tab:linear_eval}
	\footnotesize
	\begin{threeparttable}
		\begin{tabular}{l l c c c c c l}
			\toprule
			\multirow{2}{*}{\textbf{Model}} & \multirow{2}{*}{\textbf{Pretrain Dataset}} & \multirow{2}{*}{\textbf{Records}} & \multicolumn{2}{c}{\textbf{Multi-Label}} & \multicolumn{2}{c}{\textbf{Multi-Class}} & \multirow{2}{*}{\textbf{Source}} \\ \cmidrule(lr){4-5} \cmidrule(lr){6-7}
			& & & PTB-XL & CPSC2018 & PTB-XL & CPSC2018 & \\
			\midrule
			MoCo v3 & \multirow{3}{*}{CS + CODE-15} & \multirow{3}{*}{\SI{180}{K}} & - & - & 0.739 & 0.712 & \multirow{3}{*}{\cite{na2024guiding}} \\ 
			MTAE    & & & - & - & 0.807 & 0.818 & \\ 
			MLAE    & & & - & - & 0.779 & 0.794 & \\ 
			\midrule
			ST-MEM & \multirow{2}{*}{CS + CODE-15} & \multirow{2}{*}{\SI{180}{K}} & 0.882 & 0.955 & 0.879 & 0.964 & \multirow{5}{*}{\cite{kim2024learning}} \\ 
			SimCLR & & & 0.875 & 0.915 & 0.830 & 0.925 & \\ 
			ECG-FM  & UHN-ECG\tnote{4} & \SI{622}{K} & 0.878 & 0.916 & 0.856 & 0.931 & \\ 
			KED& MIMIC-IV-ECG & \SI{800}{K} & 0.885 & 0.883 & 0.888 & 0.906 & \\ 
			ECG-JEPA & CS + CODE-15 & \SI{180}{K} & 0.912 & 0.966 & 0.903 & 0.973 & \\ 
			\midrule
			JEPA\tnote{1} & \num{10} datasets\tnote{3} & \SI{1}{M}   & - & - & 0.928(03) & - & \multirow{2}{*}{\cite{weimann2025self}} \\ 
			JEPA\tnote{1} & MIMIC-IV-ECG               & \SI{800}{K} & - & - & 0.920(02) & 0.976(01) & \\ 
			\midrule
			1D-JEPA\tnote{2} & \multirow{3}{*}{CS + CODE-15} & \multirow{3}{*}{\SI{180}{K}} & 0.901(01) & 0.960(02) & 0.888(04) & 0.969(02) & \multirow{3}{*}{-} \\
			ER-JEPA          & & & 0.913(01) & 0.964(02) & 0.911(03) & 0.969(04) & \\ 
			ER-JEPA\tnote{1} & & & 0.913(00) & 0.967(01) & 0.916(02) & 0.973(01) & \\ 
			\bottomrule
		\end{tabular}
		\begin{tablenotes}
			\footnotesize
			\item[*] \textit{CS} denotes the Chapman-Shaoxing dataset.
			\item[1] \num{10} trials of downstream training.
			\item[2] Baseline 1D-JEPA model with \num{5} pretrained encoders.
			\item[3] \SI{1}{M} recordings sampled from \num{10} public datasets including CODE-15, MIMIC-IV-ECG~\cite{gow2023mimic}, and CS.
			\item[4] Private dataset from University Health Network (UHN), Toronto, Canada~\cite{mckeen2025ecg}.
		\end{tablenotes}
	\end{threeparttable}
\end{table}

The experimental results in Table~\ref{tab:linear_eval} indicate that ER-JEPA demonstrates performance improvements on the PTB-XL dataset across the board, 
with the sole exception of the highly scaled JEPA from \wei. 
The arrhythmia classification capability of the model on the CPSC2018 dataset is highly competitive with state-of-the-art models.
It is worth noting that performance differences among baseline models, including ablation studies, are comparatively marginal on the CPSC2018 dataset, where macro AUCs heavily saturate around \num{0.969}.


\subsubsection{Fine-Tuning}\label{sec:fine_tuning} 
In contrast to linear evaluation, the entire encoder is trained alongside the appended classification layer during the fine-tuning process.
For details regarding the classifier and evaluation process, refer to the \nameref{par:downstream_training} paragraph above.
The entire model was trained over \num{6} epochs using a decreasing learning rate 
with cosine decay and linear warmup~\cite{MaskedAutoencoders2021} based on a learning rate of \num{1.8e-4},
a batch size of \num{16}, and the AdamW optimizer~\cite{loshchilov2017decoupled}. 
Furthermore, data augmentations and both high- and low-pass filtering were applied, following the training procedures of ST-MEM~\cite{na2024guiding}.
\begin{table}[ht]
	\centering
	\setlength{\tabcolsep}{4.5pt}
	\caption{\textbf{Fine-Tuning Evaluation Benchmark Results.}
		Performance (macro AUC) of SSL models on downstream datasets with fine-tuning.
		Entries with multiple trials are accompanied by the standard deviation in parentheses
		(0.0xx), and the reported value is the average macro AUC over trials.
		The \textit{Source} column indicates the literature from which the benchmark scores were retrieved.
	}\label{tab:finetune}
	\footnotesize
	\begin{threeparttable}
		\begin{tabular}{l l c c c c c l}
			\toprule
			\multirow{2}{*}{\textbf{Model}} & \multirow{2}{*}{\textbf{Pretrain Dataset}} & \multirow{2}{*}{\textbf{Records}} & \multicolumn{2}{c}{\textbf{Multi-Label}} & \multicolumn{2}{c}{\textbf{Multi-Class}} & \multirow{2}{*}{\textbf{Source}} \\ \cmidrule(lr){4-5} \cmidrule(lr){6-7}
			& & & PTB-XL & CPSC2018 & PTB-XL & CPSC2018 & \\
			\midrule
			MoCo v3 & \multirow{3}{*}{CS + CODE-15} & \multirow{3}{*}{\SI{180}{K}} & - & - & 0.913 & 0.967 & \multirow{3}{*}{\cite{na2024guiding}} \\ 
			MTAE    & & & - & - & 0.910 & 0.961 & \\ 
			MLAE    & & & - & - & 0.915 & 0.973 & \\ 
			\midrule
			ST-MEM & \multirow{2}{*}{CS + CODE-15} & \multirow{2}{*}{\SI{180}{K}} & 0.929 & 0.973 & 0.910 & 0.977 & \multirow{5}{*}{\cite{kim2024learning}} \\ 
			SimCLR & & & 0.918 & 0.936 & 0.928 & 0.955 & \\ 
			ECG-FM  & UHN-ECG\tnote{4} & \SI{622}{K} & 0.899 & 0.922 & 0.895 & 0.947 & \\ 
			KED & MIMIC-IV-ECG & \SI{800}{K} & 0.901 & 0.891 & 0.906 & 0.923 & \\ 
			ECG-JEPA & CS + CODE-15 & \SI{180}{K} & 0.931 & 0.973 & 0.928 & 0.976 & \\ 
			\midrule
			JEPA\tnote{1} & \num{10} datasets\tnote{3} & \SI{1}{M}   & - & - & 0.935(02) & - & \multirow{2}{*}{\cite{weimann2025self}} \\ 
			JEPA\tnote{1} & MIMIC-IV-ECG               & \SI{800}{K} & - & - & 0.928(02) & 0.983(00) & \\ 
			\midrule
			1D-JEPA\tnote{2} & \multirow{3}{*}{CS + CODE-15} & \multirow{3}{*}{\SI{180}{K}} & 0.923(03) & 0.973(02) & 0.923(05) & 0.979(01) & \multirow{3}{*}{-} \\
			ER-JEPA          & & & 0.936(01) & 0.974(01) & 0.943(02) & 0.981(01) & \\ 
			ER-JEPA\tnote{1} & & & 0.935(01) & 0.973(01) & 0.943(03) & 0.980(01) & \\ 
			\bottomrule
		\end{tabular}
		\begin{tablenotes}
			\footnotesize
			\item[*] \textit{CS} denotes the Chapman-Shaoxing dataset.
			\item[1] \num{10} trials of downstream training.
			\item[2] Baseline 1D-JEPA model with \num{5} pretrained encoders.
			\item[3] \SI{1}{M} recordings sampled from \num{10} public datasets including CODE-15, MIMIC-IV-ECG~\cite{gow2023mimic}, and CS.
			\item[4] Private dataset from University Health Network (UHN), Toronto, Canada~\cite{mckeen2025ecg}.
		\end{tablenotes}
	\end{threeparttable}
\end{table}

Table~\ref{tab:finetune} presents the benchmark results for the fine-tuning evaluation.
As with the linear evaluation, ER-JEPA excels on the PTB-XL dataset, outperforming the previous state-of-the-art,
and attains highly competitive state-of-the-art performance for arrhythmia classification on CPSC2018.

Notably, there is a significant increase in the macro AUC scores compared to the linear evaluation, in contrast to the modest \num{0.007}--\num{0.008} AUC increase reported for the JEPA from \wei~\cite{weimann2025self}.
This substantial increase following fine-tuning suggests that while the pretraining process may not inherently force 
the encoder to learn features perfectly aligned for a specific downstream task,
the resulting hierarchical embeddings are highly adaptable, allowing the encoder to rapidly tailor its representations during the fine-tuning phase.

\subsection{Computational Efficiency}\label{sub:computational_efficiency} 
Most ViT-based SSL models in the benchmark, 
such as the encoders proposed by Weimann and Conrad~\cite{weimann2025self},
ST-MEM~\cite{na2024guiding},
and ECG-JEPA, 
share common hyperparameter standards.
Following the configuration of the original ViT~\cite{dosovitskiy2020image} or 
the foundational Transformer~\cite{vaswani2017attention}, 
an embedding dimension equal to a multiple of \num{768} and an encoder depth of \num{12} are prevalent baseline choices.
ER-JEPA also follows this general convention,
utilizing an embedding dimension of $\demb = \num{384}$,
but splits the network depth, equally distributing six attention layers to the channel encoder and six to the temporal encoder.

While sections~\ref{sec:linear_evaluation} and \ref{sec:fine_tuning} prove
that this configuration, $\demb=\num{384}$ with depths $(6,6)$, is highly effective,
the ER-JEPA encoder also gains a significant advantage
with respect to resource usage,
as half of its attention layers (the temporal encoder) process purely temporal tokens 
without requiring expensive multichannel operations.
The multichannel processing is dedicated entirely to the channel encoder, 
which itself achieves efficiency through a strictly focused scope.
Compared to the temporal encoder, the channel encoder processes a much smaller sequence length of $N=\numch$, at the expense of increasing the effective batch size
by a factor of $\numt$.
However, in the current implementation ($\numch = 8$ and $\numt = 50$),
this increase in effective batch size dominates the computational cost over the reduced sequence length, despite the $\mathcal{O}(N^2)$ reduction in attention complexity.
Furthermore, computational overhead caused by tensor handling in the PyTorch implementation widens the execution time gap between the temporal and channel encoders, regardless of the configuration.

\begin{figure}[htbp]
	\begin{center}
		\captionsetup[subfigure]{justification=centering}
		\begin{subfigure}{0.95\textwidth}
			\includegraphics[width=\textwidth]{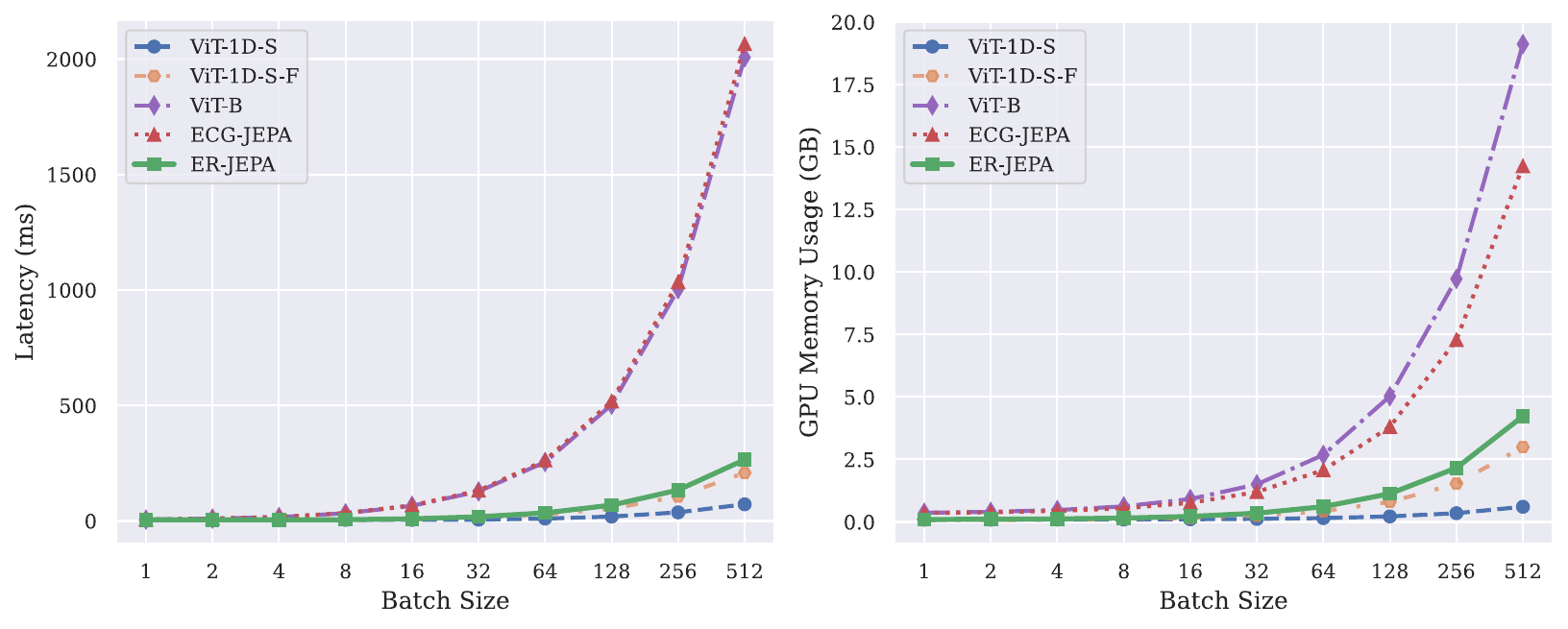}
			\caption{Classification Benchmark Encoders}\label{subfig:speed_models_benchmark}
		\end{subfigure}
		\begin{subfigure}{0.95\textwidth}
			\includegraphics[width=\textwidth]{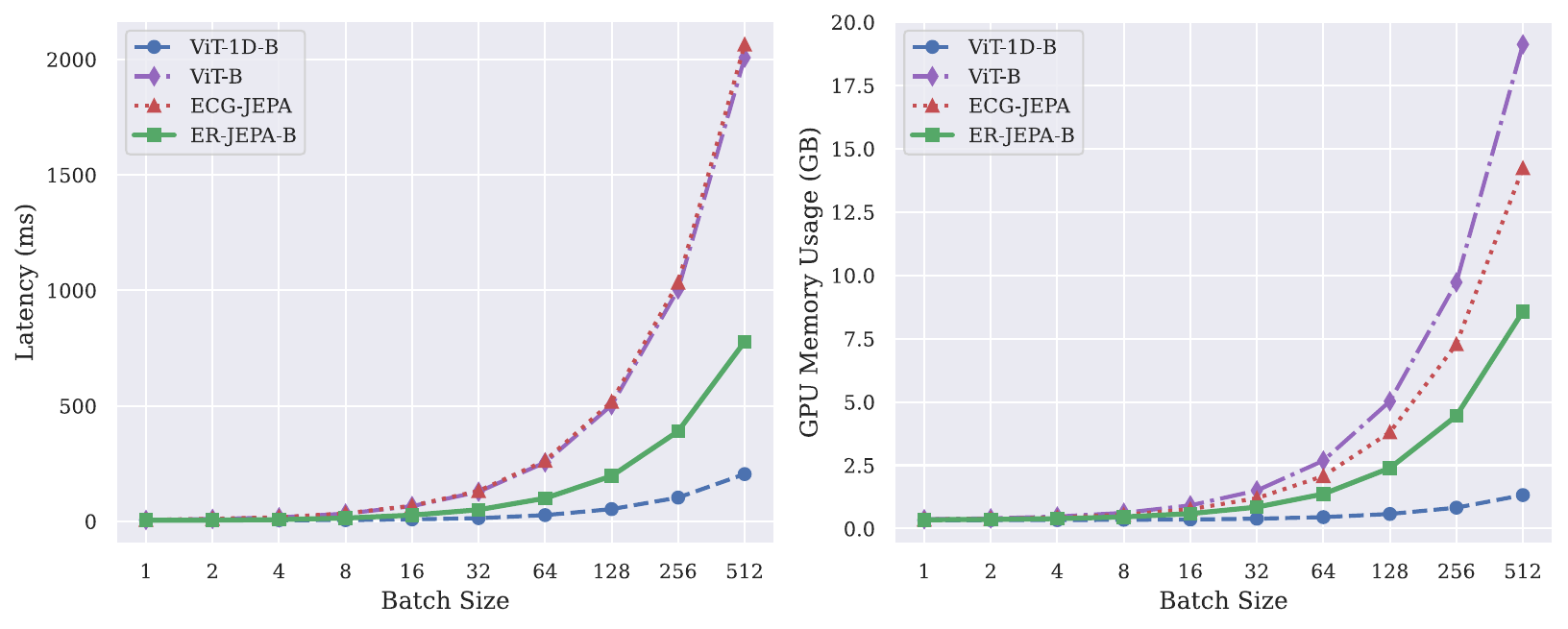}
			\caption{Encoders under Unified Embedding Dimension}\label{subfig:speed_models_768}
		\end{subfigure}
	\end{center}
	\caption{
		\textbf{Computational Efficiency Comparison.}
		The ViT-based encoders are evaluated based on batch latency and peak GPU memory usage.
		(a) Performance utilizing the native hyperparameter settings from the downstream classification
		benchmark.
		(b) A controlled comparison where all encoders are standardized to a unified embedding dimension
		of $\demb = 768$.
	}\label{fig:speed_comparison}
\end{figure}

Figure~\ref{fig:speed_comparison}
presents an efficiency comparison among the ViT encoders of various SSL models.
The ER-JEPA encoder is compared in two settings: (a) under the native settings used in the downstream classification benchmark, and (b) under a unified embedding dimension of $\demb = 768$.
The compared models include a 1D ViT-S; a 2D ViT-B, representing the encoder of ST-MEM; and the Cross-Pattern ViT from ECG-JEPA~\cite{kim2024learning}.
Furthermore, ViT-1D-S-F, a closer surrogate for the encoder in \wei,
is included in the native setting benchmark 
to account for their specific data resolution (\SI{500}{Hz} data with a smaller patch size of $p=25$).

With the hyperparameters used in the classification benchmark, the ER-JEPA encoder is the fastest among
the architectures that explicitly process multichannel tokens, achieving speeds relatively close to the baseline ViT-1D encoder (which entirely bypasses multichannel attention).
The ER-JEPA encoder achieves up to a \num{7.8}$\times$ speedup under benchmark conditions,
and maintains a maximum \num{2.6}$\times$ speedup even under the unified embedding dimension of $\demb=768$.

The memory allocation exhibits a trend similar to the speed efficiency.
Compared to the other multichannel ViT encoders in the benchmark, ER-JEPA
requires less than one-third of the memory for small batch sizes, and less than one-fourth for larger batch sizes.
When all models are scaled to an embedding dimension of $\demb = 768$, memory usage is similar for a unit batch size, but the baseline multichannel encoders scale to consume between \num{1.5} and \num{2} times as much memory as ER-JEPA for larger batch sizes.

While the basic 1D ViT achieves the lightest computational footprint by delegating
multichannel analysis entirely to the tokenization step,
the finer-resolution surrogate (ViT-1D-S-F) demands significantly heavier processing.
As a result, the efficiency gap between ViT-1D-S-F and the ER-JEPA encoder narrows, with ER-JEPA requiring only a \SI{25}{\percent} increase in operation time and \SI{42}{\percent} more memory despite multichannel processing.

\paragraph{Hardware and Software Specifications}\label{par:hardware_and_software_specifications} 
The computational benchmark was executed on a workstation
equipped with an Intel® Core™ i9-10940X processor and
an NVIDIA GeForce RTX 3090 GPU (\SI{24}{GB} VRAM),
utilizing Python 3.12.3, PyTorch 2.10.0, and CUDA 12.8 running on Ubuntu 20.04.6 LTS.

\paragraph{Computation Benchmark Setup}\label{par:computation_benchmark_setup} 
In the experimental script, a random tensor of shape $(B,8,2500)$ was utilized as a single batch of recordings. This was partitioned into a shape of $(B,8,50,50)$, or $(B,50,50)$ for the 1D-JEPA, using a unified patch size of $p=50$.
For the ViT-1D-S-F model, which operates on a different data resolution, an input tensor of shape $(B,8,5000)$ was used with a patch size of $p=25$, yielding an initial patch tensor of shape $(B,200,25)$.
The latency per batch was averaged over \num{100} iterations, 
preceded by a warmup phase of the same number of iterations.
After the warmup phase, memory usage was cleared utilizing 
\code{torch.cuda.reset\_peak\_memory\_stats},
and both phases were followed by a \code{torch.cuda.synchronize} call. 
For latency measurement, \code{time.perf\_counter} recorded execution times with high precision.
Just-In-Time (JIT) compilation using \code{torch.compile} was disabled during 
the experiment.
Each timing experiment was repeated five times; the standard deviations reached a maximum of \SI{1}{\percent} for small batch sizes (under four), while the majority remained well below \SI{1}{\percent}.

\subsection{Qualitative Assessment of Intermediate Representations}\label{sub:qualitative_assessment_of_intermediate_representations} 
In this section, we qualitatively assess the characteristics of the representations produced by the channel encoder.
Because the representation from the channel encoder corresponds to a specific time interval, as with the output tokens of the temporal encoder, 
both representations can be utilized in downstream applications involving time-interval analysis.
To this end, we compare the efficacy of each representation 
in an ECG segmentation task.
We restrict this analysis strictly to a qualitative proof-of-concept, reserving comprehensive quantitative evaluation for future studies.

\subsubsection{Delineation of Key ECG Waveforms}\label{sec:delineation_of_key_ecg_waveforms} 
The three primary waveforms of an ECG correspond 
to distinct events in the cardiac cycle: 
the P-wave represents atrial depolarization, 
the QRS complex indicates ventricular depolarization, 
and the T-wave signifies ventricular repolarization.
Because each waveform is a projection of a corresponding cardiac event, any functional deviation of the heart is directly reflected in its morphology.
While the accurate delineation of these key waveforms is a crucial step in clinical diagnosis, it is also an absolute prerequisite for automated ECG analysis such as large-scale statistical filtering.

\paragraph{Difficulty of Delineation}\label{par:difficulty_of_delineation} 
Despite its importance, the scarcity of comprehensively annotated datasets and the morphological variance across different arrhythmias intensify the difficulty of the delineation task~\cite{joung2024deep}.
For instance, an obstruction or delay in electrical conduction between 
the sinus node and the Purkinje fibers is referred to as an 
Atrioventricular (AV) block, a condition with varying degrees of severity~\cite{thaler2021only}.
AV blocks pose significant challenges for automated delineation due to the potential overlap 
between delayed P-waves and other waveforms, the complete absence of a subsequent ventricular response, or the disassociation of atrial and ventricular activity.

\subsubsection{ECG Segmentation}\label{sec:ecg_segmentation} 
Building on the U-Net architecture~\cite{ronneberger2015u} and its variants~\cite{zhou2018unet++,huang2020unet}, the work of Joung et al.~\cite{joung2024deep} demonstrated the successful application of deep learning frameworks for ECG delineation across diverse arrhythmias.
The essence of U-Net-based architectures is a symmetric encoding-decoding structure,
where each module forms half of the `U' shape.
The layers in the encoding path progressively compress the preceding signal into a condensed form,
reaching maximum compression at the bottleneck, where each temporal step encompasses a broad receptive field of the input time series.

The intermediate representations of the ER-JEPA channel encoder act as representative tokens for corresponding time intervals, capturing the mutual correlation between concurrent patches across different leads. 
Subsequently, the output tokens of the temporal encoder are derived by applying self-attention
to these intermediate representations.
Hence, the sequence of either representation across all time intervals can be interpreted as
an encoded signal, compressed by a factor of the patch size $p$, with the embedding dimension acting as the feature map depth.
Based on this interpretation, either encoder of ER-JEPA can 
replace the standard encoding path of a U-Net-style segmentation network.
The following demonstration utilizes a segmentation network designed from this premise. 
Based on UNet 3+~\cite{huang2020unet,joung2024deep}, the modified model retains only the full-scale skip connections of the decoding layers,
as the standard encoding path is replaced by the ER-JEPA encoder.
This design serves merely as a proof-of-concept;
architectural exploration and further hyperparameter optimization are reserved for future work.

\subsubsection{Over-Smoothing}\label{sec:over_smoothing} 
While the integration of attention mechanisms with SSL has significantly advanced representation learning,
structural designs must be carefully calibrated depending on the target task.
Whether due to the inherent nature of the self-attention algorithm~\cite{dong2021attention}
or specific structural learning designs~\cite{simeoni2025dinov3},
transformer architectures are prone to the convergence of every token in a sequence into a 
uniform vector, a phenomenon referred to as \textbf{over-smoothing}, \textbf{token uniformity}, or \textbf{rank collapse}.
Although standard transformer implementations 
include measures to mitigate this over-smoothing~\cite{dong2021attention},
tasks requiring fine-grained, localized inference, such as segmentation, demand extra structural care.
Along with various architectural solutions, utilizing tokens from intermediate layers is known to facilitate
token diversity and prevent the degradation of high-frequency signals~\cite{lee2026frequency}.

In ER-JEPA, because the representation of the channel encoder exclusively
captures mutual relations within concurrent patches,
the reconstructed representations of time intervals remain mutually independent
prior to temporal processing.
From this perspective, the channel encoder representations feature
fine-grained, distinctive signals free from global influence.
Despite the absence of global inference, it is theoretically advantageous for facilitating localized inference in segmentation tasks.
\subsubsection{Case Study: P-Wave Delineation in Complex Arrhythmias}\label{sec:case_study_p_wave_delineation_in_complex_arrhythmias} 
Restricting our focus strictly to qualitative evaluation,
we note the potential capability of ER-JEPA in the ECG segmentation task and provide
a selected example illustrating the theoretical advantage of the channel encoder representation for local feature extraction.
Each segmentation model was fine-tuned under an identical environment utilizing a
filtered subset of the Shaoxing dataset.
The segmentation labels were primarily generated by the segmentation model of Joung et al.~\cite{joung2024deep}.
Labels for \num{42} second-degree AV block samples and \num{24} third-degree AV block samples in the Shaoxing dataset were manually annotated.
Approximately \num{1000} recordings, including \num{66} high-degree AV block samples,
were used in total for the segmentation fine-tuning.

\begin{figure}[htbp]
	\begin{center}
		\includegraphics[width=0.7\textwidth]{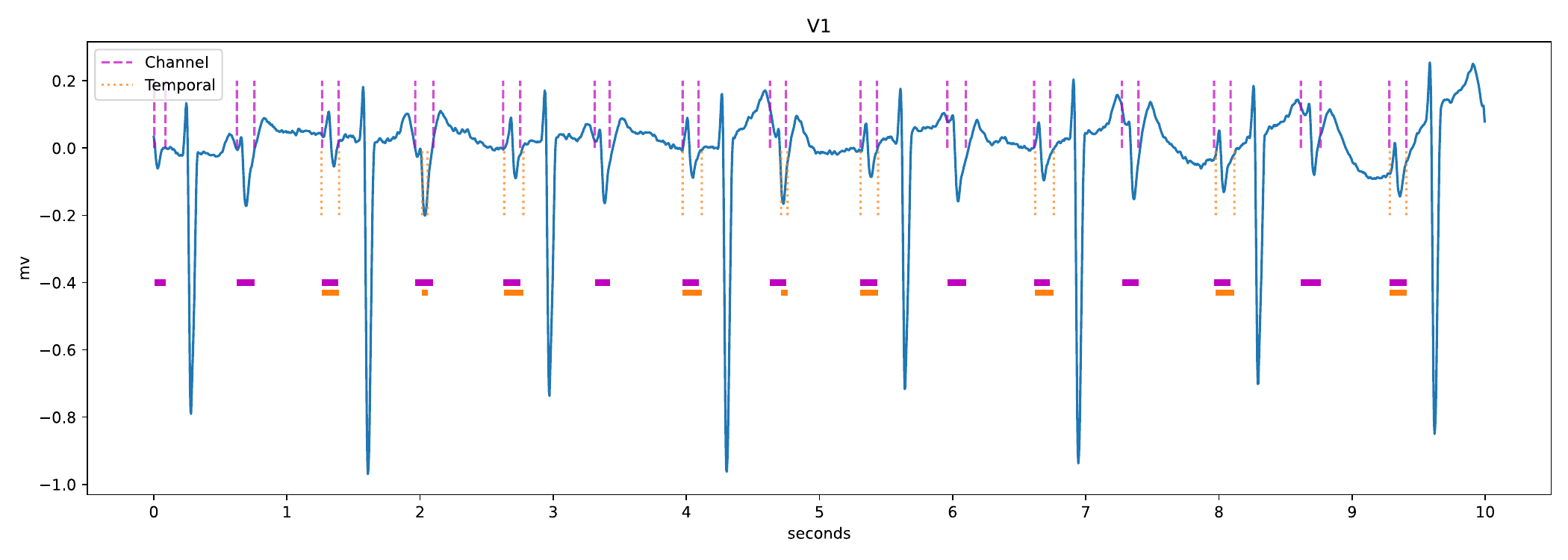}
	\end{center}
	\caption{\textbf{Qualitative Comparison of P-Wave Delineation.}
		A selected recording (Shaoxing dataset record ID JS04915)
		with a 2nd-degree Atrioventricular Block (AVB2) serves as an exemplary sample to
		illustrate the comparison of local feature extraction performance between intermediate (Channel)
		and final (Temporal) representations.
		The delineations of each segmentation model
		are depicted on the recording (Lead V1).
		Despite absence of subsequent QRS complexes and the overlapping of P-waves and T-waves due to conduction delay,
		the channel encoder segmentation model successfully delineates every P-wave in the recording.
		The full model detects the presence of the buried P-waves, but the signal is not strong enough.
	}\label{fig:segmentation-result}
\end{figure}

Figure~\ref{fig:segmentation-result} presents the segmentation result
of a selected example delineated by two segmentation models, 
each featuring either the isolated channel encoder or the entire ER-JEPA encoder, paired with a UNet 3+-based decoder add-on.
For a qualitative evaluation of the differences between the representations
at different stages,
the first model utilized only the channel encoder, while
the second model utilized the final encoding of the entire ER-JEPA encoder as the input
to the decoder.
Although this selected case is illustrative rather than exhaustive
(especially without an intensive study, which is reserved for future work),
it provides a clear visualization of the hypothesis discussed in
Section~\ref{sec:over_smoothing}.

The visualization features a successful case of P-wave delineation by the channel encoder segmentation model on a sample with a 2nd-degree AV block.
Here, the channel encoder segmentation model locates every non-conducted P-wave that is
overlapped with a T-wave.
The full model detects the presence of buried P-waves in a few cases, 
but the signal strength is insufficient in this specific sample.
While the full model successfully detected buried P-waves in some other samples, this example
illustrates the potential signal suppression caused by a heavy reliance on the global rhythm
acquired through temporal inference.

In summary, we have explored the potential capability of ER-JEPA in ECG
segmentation and provided an exemplary sample that illustrates our hypothesis
regarding the characteristics of intermediate representations, specifically in relation to the
over-smoothing of self-attention operations.
This targeted demonstration was successful even without extensive
structural design and optimization of the segmentation network;
however, a comprehensive quantitative evaluation 
and an examination of optimized architectures are reserved for future work.



\section{Ablations}\label{sec:ablations} 
The assessment for the ablation studies follows the same downstream evaluation procedure outlined in Section~\ref{par:downstream_training}.
The subsequent sections feature ablation studies evaluating various hyperparameters based on linear probing and fine-tuning performance.
While each analysis demonstrates that the default hyperparameter configuration is optimal,
fine-tuning evaluations exhibit significantly narrower performance margins,
demonstrating highly stable performance across diverse configurations.

\subsection{Linear Evaluation Across Ablations}\label{sec:linear_evaluation_across_ablations} 

\paragraph{Masking Strategy}\label{par:masking_strategy} 
As discussed in Section~\ref{sec:masking_strategy}, two masking strategies for the temporal pretraining process
are compared in Table~\ref{tab:mask_strategy}.
To leverage the effects of both strategies, a `mixed' scheme was also tested,
which applies a block mask to half of the batch and a sparse mask to the remainder.
Table~\ref{tab:mask_strategy} indicates that while every mask scheme demonstrates strong performance,
the block mask scheme outperforms the others on the PTB-XL classification tasks.
\begin{table}[!ht]
    \centering
    \caption{\textbf{Linear Evaluation Across Masking Strategy.}}\label{tab:mask_strategy}
    \begin{tabular}{l c c c c}
        \toprule
        \multirow{2}{*}{\textbf{Strategy}} & \multicolumn{2}{c}{\textbf{Multi-Label}} & \multicolumn{2}{c}{\textbf{Multi-Class}} \\ \cmidrule(lr){2-3} \cmidrule(lr){4-5}
        & PTB-XL & CPSC2018 & PTB-XL & CPSC2018 \\
        \midrule
        Sparse & 0.909(02) & 0.962(03) & 0.903(05) & 0.968(01) \\
        Mix    & 0.909(02) & 0.965(02) & 0.903(04) & 0.970(03) \\
        \midrule
        Block  & 0.913(01) & 0.964(02) & 0.911(03) & 0.969(04) \\
        \bottomrule
    \end{tabular}
\end{table}


\paragraph{Mask Size}\label{par:mask_size} 
Considering the semantic learning objectives of JEPA pretraining,
the size of the context interval is a crucial factor.
If the context is too wide, the prediction task becomes comparatively trivial,
whereas a critically small context may hinder learning by making the task excessively difficult.
A training strategy utilizing smaller masks sampled from the range of \num{1.4}--\SI{2.0}{seconds}
outperformed a larger mask scheme sampled from the range of \num{2.8}--\SI{3.6}{seconds}.
However, the broader range of \num{1.4}--\SI{3.6}{seconds}, which encompasses both, proved optimal
by preventing the model from overfitting to a fixed trend in size sampling.
\begin{table}[!ht]
    \centering
    \caption{\textbf{Linear Evaluation Across Mask Size.}}\label{tab:mask_size}
    \begin{tabular}{l c c c c}
        \toprule
        \multirow{2}{*}{\textbf{Mask Size}} & \multicolumn{2}{c}{\textbf{Multi-Label}} & \multicolumn{2}{c}{\textbf{Multi-Class}} \\ \cmidrule(lr){2-3} \cmidrule(lr){4-5}
        & PTB-XL & CPSC2018 & PTB-XL & CPSC2018 \\
        \midrule
        Large & 0.910(01) & 0.961(03) & 0.903(03) & 0.965(03) \\
        Small & 0.911(01) & 0.963(03) & 0.907(04) & 0.968(03) \\
        \midrule
        Both  & 0.913(01) & 0.964(02) & 0.911(03) & 0.969(04) \\
        \bottomrule
    \end{tabular}
\end{table}


\paragraph{Batch Size}\label{par:batch_size} 
For the current pretraining dataset, varying the batch size yielded
the linear evaluation performances detailed in Table~\ref{tab:batch_size}.
The variation in benchmark performance across these batch sizes is the least significant among all the ablation studies.
A batch size of \num{64} is marginally optimal from an overall perspective; however,
optimal performance is not guaranteed under different data resolutions, pretraining dataset sizes,
embedding dimensions, or layer compositions.
\begin{table}[!ht]
    \centering
    \caption{\textbf{Linear Evaluation Across Batch Size.}}\label{tab:batch_size}
    \begin{tabular}{l c c c c}
        \toprule
        \multirow{2}{*}{\textbf{Batch Size}} & \multicolumn{2}{c}{\textbf{Multi-Label}} & \multicolumn{2}{c}{\textbf{Multi-Class}} \\ \cmidrule(lr){2-3} \cmidrule(lr){4-5}
        & PTB-XL & CPSC2018 & PTB-XL & CPSC2018 \\
        \midrule
        32  & 0.913(02) & 0.962(02) & 0.907(03) & 0.970(02) \\
        128 & 0.912(00) & 0.964(02) & 0.909(03) & 0.970(03) \\
        \midrule
        64  & 0.913(01) & 0.964(02) & 0.911(03) & 0.969(04) \\
        \bottomrule
    \end{tabular}
\end{table}


\paragraph{Time Range for Channel JEPA Training}\label{par:time_range_for_channel_jepa_training} 
Because ER-JEPA consists of two JEPA modules, it involves two distinct pretraining processes: 
one predicting target channels from a given context, and another predicting target time intervals from a given context.
Because the pretraining of the channel JEPA is not restricted by the subsequent temporal pretraining,
the range of the time interval utilized for channel pretraining can be freely selected.
Table~\ref{tab:time-range} compares the linear evaluation results of models trained with different time interval selections for 
channel pretraining.
Specifically, a model with the channel JEPA pretrained exclusively over the context time interval of the temporal pretraining is compared against
models pretrained over all available time intervals and over the target time interval of the temporal pretraining.
The selected time range for channel pretraining affects the the overall scope of data utilized during the pretraining process.
\begin{table}[!ht]
    \centering
    \caption{\textbf{Linear Evaluation Across Training Time Interval.}}\label{tab:time-range}
    \begin{tabular}{l c c c c}
        \toprule
        \multirow{2}{*}{\textbf{Training Interval}} & \multicolumn{2}{c}{\textbf{Multi-Label}} & \multicolumn{2}{c}{\textbf{Multi-Class}} \\ \cmidrule(lr){2-3} \cmidrule(lr){4-5}
        & PTB-XL & CPSC2018 & PTB-XL & CPSC2018 \\
        \midrule
        Target Mask     & 0.913(02) & 0.966(03) & 0.906(01) & 0.969(03) \\
        Random Sampling & 0.911(03) & 0.967(00) & 0.907(05) & 0.971(03) \\
        All             & 0.912(01) & 0.966(02) & 0.908(04) & 0.970(05) \\
        \midrule
        Context Mask    & 0.913(01) & 0.964(02) & 0.911(03) & 0.969(04) \\
        \bottomrule
    \end{tabular}
\end{table}


\paragraph{Embedding Dimension}\label{par:embedding_dimension} 
Similar to the issues discussed regarding the batch size ablation and 
the sensitivity of hierarchical representation pretraining,
a model with a larger embedding dimension of $\demb=768$ may require a significantly different pretraining environment
to achieve optimal performance.
However, the ablation study demonstrates that
a smaller embedding dimension ($\demb=384$) outperforms the larger model within the current, controlled environment,
a trend similarly observed in the JEPA proposed by \wei~\cite{weimann2025self}.
\begin{table}[!ht]
    \centering
    \caption{\textbf{Linear Evaluation by Embedding Dimension.}}\label{tab:dimension}
    \begin{tabular}{l c c c c}
        \toprule
        \multirow{2}{*}{\textbf{Embedding Dimension}} & \multicolumn{2}{c}{\textbf{Multi-Label}} & \multicolumn{2}{c}{\textbf{Multi-Class}} \\ \cmidrule(lr){2-3} \cmidrule(lr){4-5}
        & PTB-XL & CPSC2018 & PTB-XL & CPSC2018 \\
        \midrule
        768 & 0.908(04) & 0.951(05) & 0.894(04) & 0.953(04) \\
        \midrule
        384 & 0.913(01) & 0.964(02) & 0.911(03) & 0.969(04) \\
        \bottomrule
    \end{tabular}
\end{table}


\paragraph{Without Channel JEPA}\label{par:without_channel_jepa} 
As Table~\ref{tab:misc} indicates, pretraining solely with the temporal JEPA
produced a negative effect on representation learning (see Section~\ref{par:comparison_without_channel_jepa}).
By exhibiting definitively inferior performance compared to the complete model, this ablation provides strong empirical evidence for the structural necessity of the channel JEPA.

\paragraph{Large Loss and Dropout Layer}\label{par:large_loss} 
Within the current pretraining environment, repeated trials rarely produced 
significantly underperforming models (see Section~\ref{par:loss_drop}).
Table~\ref{tab:misc} presents the downstream results of these anomalous models, which are characterized by a loss value at Epoch 2 larger than \num{0.008}.
Additionally, Table~\ref{tab:misc} highlights the notable performance decrease observed when the dropout layer is completely omitted from the architecture.
\begin{table}[!ht]
    \centering
    \caption{\textbf{Linear Evaluation Across Miscellaneous Ablations.}}\label{tab:misc}
    \begin{tabular}{l c c c c}
        \toprule
        \multirow{2}{*}{\textbf{Ablations}} & \multicolumn{2}{c}{\textbf{Multi-Label}} & \multicolumn{2}{c}{\textbf{Multi-Class}} \\ \cmidrule(lr){2-3} \cmidrule(lr){4-5}
        & PTB-XL & CPSC2018 & PTB-XL & CPSC2018 \\
        \midrule
        No Dropout & 0.908(01) & 0.961(06) & 0.903(02) & 0.969(02) \\
        No Channel & 0.902(03) & 0.959(06) & 0.892(07) & 0.965(06) \\
        Large Loss & 0.904(03) & 0.953(12) & 0.900(03) & 0.954(10) \\
        \midrule
        Normal     & 0.913(01) & 0.964(02) & 0.911(03) & 0.969(04) \\
        \bottomrule
    \end{tabular}
\end{table}


\subsection{Fine-Tuning Across Ablations}\label{sec:finetuning_across_ablations} 
In contrast to the linear evaluation, the performance across ablation studies on the fine-tuning downstream task
did not exhibit significant deviation, although 
the default hyperparameters remained optimal by a minimal margin.
Table~\ref{tab:ablation_finetune} presents the average and standard deviation of
the benchmark results grouped by broader categories.
While most of the categories showed heavily saturated performance,
the encoders trained without the channel JEPA noticeably and consistently deviated from the majority.
\begin{table}[!ht]
	\centering
	\caption{\textbf{Fine-Tuning Performance Across Ablation Studies.}
		To summarize the results concisely, several minor variations are grouped into broader categories (excluding the default configuration):
		`Mask All' encompasses all mask size and masking strategy ablations;
		`Interval All' aggregates the variations in channel training intervals; and
		`Batch All' includes all tested batch sizes.
	}\label{tab:ablation_finetune}
	\begin{tabular}{l c c c c}
		\toprule
		\multirow{2}{*}{\textbf{Category}} & \multicolumn{2}{c}{\textbf{Multi-Label}} & \multicolumn{2}{c}{\textbf{Multi-Class}} \\ \cmidrule(lr){2-3} \cmidrule(lr){4-5}
		& PTB-XL & CPSC2018 & PTB-XL & CPSC2018 \\
		\midrule
		No Channel   & 0.931(01) & 0.975(02) & 0.936(03) & 0.980(01) \\ 
		Large Loss   & 0.934(00) & 0.974(03) & 0.943(01) & 0.979(02) \\ 
		No Drop Rate & 0.936(01) & 0.973(02) & 0.941(5)  & 0.980(01) \\ 
		$\demb =768$ & 0.935(01) & 0.974(01) & 0.938(5)  & 0.978(00) \\ 
		Mask All     & 0.935(01) & 0.974(01) & 0.943(02) & 0.980(01) \\ 
		Interval All & 0.936(01) & 0.974(01) & 0.943(02) & 0.980(01) \\ 
		Batch All    & 0.936(01) & 0.975(01) & 0.942(03) & 0.981(01) \\ 
		\midrule
		Default      & 0.936(01) & 0.974(01) & 0.943(02) & 0.981(01) \\ 
		\bottomrule
	\end{tabular}
\end{table}


\section{Conclusion}\label{sec:conclusion} 
This study demonstrates the effectiveness of a lightweight framework 
utilizing hierarchical joint-representation learning on ECG data.
Our empirical results reveal that, contrary to expectations regarding structural sensitivity, 
the concatenation of JEPAs proves robust against representation collapse. 
Furthermore, this dual-JEPA approach facilitates enhanced representation learning compared to an identical architecture relying on a single JEPA.
By adopting a two-stage structure that 
encodes each multichannel time interval into a holistic embedding before processing them as a univariate time series,
ER-JEPA achieves highly efficient performance
characterized by reduced memory usage and rapid computational speed.
Validated through extensive downstream assessments,
the hierarchical structure of ER-JEPA demonstrates highly competitive performance
across standard benchmarks, notably surpassing existing state-of-the-art models on the PTB-XL multi-class classification task with a macro AUC of \num{0.943}.

\bibliographystyle{unsrt}
\bibliography{main.bib}

\newpage
\appendix
\section{Hyperparameters}\label{sec:hyperparameters} 
In this appendix, we provide the detailed hyperparameters used for the pretraining, 
linear evaluation, and fine-tuning phases of the ER-JEPA model.

\subsection{Model Architecture}\label{sub:model_architecture}
The ER-JEPA architecture comprises a channel and a temporal module, each with a Vision Transformer (ViT) backbone.
Most configurations are inherited from widely used baseline models, 
with a slight modification to the network depth, distributing \num{12} layers evenly across the two encoders.
Because the channel JEPA processes tokens indexed by (channel, interval), it utilizes 2D sinusoidal positional encoding~\cite{MaskedAutoencoders2021}. 
Conversely, the temporal JEPA processes purely temporal tokens and utilizes standard 1D sinusoidal positional encoding~\cite{vaswani2017attention}.
After every attention layer and MLP, a dropout layer with a probability of \num{0.1} is applied.

\begin{table}[!ht]
    \centering
    \caption{\textbf{Architecture Hyperparameters.}}\label{tab:arch_hyperparams}
    \begin{tabular}{l l}
    \toprule
    \textbf{Hyperparameter} & \textbf{Value} \\ 
    \midrule
    Embedding Dimension ($\demb$) & \num{384} \\ 
    Encoder Depth (Channel \& Temporal) & \num{6} \\ 
    Encoder Heads (Channel \& Temporal) & \num{12} \\ 
    Predictor Embedding Dimension & \num{192} \\ 
    Predictor Depth (Channel \& Temporal) & \num{3} \\ 
    Predictor Heads (Channel \& Temporal) & \num{6} \\ 
    MLP Expansion Ratio & $\times$\num{4} \\ 
    Input Sampling Frequency & \SI{250}{Hz} \\ 
    Patch Size ($p$) & \num{50} \\ 
    Positional Encoding (2D \& 1D) & Sinusoidal  \\ 
    Drop Path Rate (Between Layers) & \num{0.1} \\ 
    \bottomrule
    \end{tabular}
\end{table}

\subsection{Pretraining}\label{sub:pretraining_hyperparameters}
Table~\ref{tab:pretrain_hyperparams} summarizes the configurations utilized during the ER-JEPA pretraining phase.
Following standard practice, a weight decay of \num{0.05} was applied exclusively to multidimensional weight parameters, while biases and one-dimensional normalization parameters were excluded from weight decay.
Data augmentations, including baseline wander, Gaussian noise, and powerline noise, were applied during pretraining.

\begin{table}[!ht]
    \centering
    \caption{\textbf{Pretraining Hyperparameters.}}\label{tab:pretrain_hyperparams}
\begin{tabular}{l l}
    \toprule
    \textbf{Hyperparameter} & \textbf{Configuration} \\
    \midrule
    Epochs & \num{300} \\ 
    Batch Size & \num{64} \\ 
    Optimizer & AdamW \\ 
    Scheduler & SGDR \\ 
    Base Learning Rate & \num{2e-4} \\ 
    Minimum Learning Rate & \num{5e-5} \\ 
    Warmup Epochs & \SI{10}{\percent} \\ 
    SGDR Restarts & \num{8} \\ 
    Weight Decay & \num{0.05} \\ 
    Target EMA Momentum & \num{0.996} $\rightarrow$ \num{1.0} \\ 
    Number of Context Channels & \num{2}--\num{4} \\ 
    Number of Target Channels & \num{2}--\num{3} \\ 
    Number of Channel Target Masks & \num{2} \\ 
    Channel Sampling Weight & I, II: 3, V1--V6: 1 \\ 
    Context Mask Size Range & \num{1.4}--\SI{3.6}{s} \\ 
    Target Mask Size Range & \num{0.8}--\SI{1.4}{s} \\ 
    Number of Temporal Target Masks & \num{4} \\ 
    Masking Strategy & Block \\ 
    Channel Time Interval & Temporal Context \\ 
    \bottomrule
    \end{tabular}
\end{table}

\subsection{Downstream Tasks}\label{sub:downstream_hyperparameters}
Table~\ref{tab:downstream_hyperparams} details the hyperparameters used for the downstream linear evaluation and fine-tuning tasks on the PTB-XL and CPSC2018 datasets.
The base evaluation procedure follows the methodology outlined in Section~\ref{par:downstream_training}, incorporating additional data augmentations and filtering during the fine-tuning phase, as described in Section~\ref{sec:fine_tuning}.

\begin{table}[!ht]
    \centering
    \caption{\textbf{Downstream Task Hyperparameters.}}\label{tab:downstream_hyperparams}
    \begin{tabular}{l l l}
    \toprule
    \textbf{Hyperparameter} & \textbf{Linear Evaluation} & \textbf{Fine-Tuning} \\ 
    \midrule
    Epochs & \num{70} & \num{6} \\ 
    Batch Size & \num{16} & \num{16} \\ 
    Scheduler & Cosine Decay & Cosine Decay with Warmup \\ 
    Optimizer & AdamW & AdamW \\ 
    Base Learning Rate & \num{5e-3} & \num{1.8e-4} \\ 
    Minimum Learning Rate & \num{0} & \num{0} \\ 
    Weight Decay & \num{0.01} & \num{0.05} \\ 
    \bottomrule
    \end{tabular}
\end{table}

\end{document}